\documentclass[lettersize,journal]{IEEEtran}

\usepackage{amsmath,amsfonts}
\usepackage{algorithm}
\usepackage{algorithmic}
\usepackage{array}
\usepackage{textcomp}
\usepackage{verbatim}
\usepackage{graphicx}
\def\BibTeX{{\rm B\kern-.05em{\sc i\kern-.025em b}\kern-.08em
    T\kern-.1667em\lower.7ex\hbox{E}\kern-.125emX}}
\usepackage{balance}
\usepackage{cite}
\usepackage{subfigure}
\usepackage{multirow}
\usepackage[table]{xcolor}
\usepackage{amssymb}
\usepackage{footmisc}

\usepackage{url}

\begin{document}
\title{EmoBipedNav: Emotion-aware Social Navigation for Bipedal Robots with Deep Reinforcement Learning}

\author{Wei Zhu$^1$, Abirath Raju$^1$, Abdulaziz Shamsah$^2$, Anqi Wu$^3$, Seth Hutchinson$^4$, and Ye Zhao$^1$
\thanks{This work was supported in part by the Office of Naval Research (ONR) under Grant N000142312223, in part by the National Science Foundation (NSF) under Grant $\#$IIS-1924978, Grant $\#$CMMI-2144309, and Grant $\#$FRR-2328254, and in part by the United States Department of Agriculture (USDA) under Grant 2023-67021- 41397. (\textit{Corresponding author: Ye Zhao}.)}
\thanks{$^1$The authors are with the Laboratory for Intelligent Decision and Autonomous Robots, Woodruff School of Mechanical Engineering, Georgia Institute of Technology, Atlanta, GA 30313, USA. {\tt\small \{wzhu328, araju60, yezhao\}@gatech.edu}}
\thanks{$^2$College of Engineering and Petroleum, Kuwait University, PO Box 5969, Safat, 13060, Kuwait. {\tt\small abdulaziz.shamsah@ku.edu.kw}}
\thanks{$^3$School of Computational Science and Engineering, Georgia Institute of Technology, Atlanta, GA, 30332 USA. {\tt\small anqiwu@gatech.edu}}
\thanks{$^4$Khoury College of Computer Sciences, Northeastern University, Boston, MA 02115, USA. {\tt\small s.hutchinson@northeastern.edu}}}


\maketitle

\begin{abstract}
This study presents an emotion-aware navigation framework -- EmoBipedNav -- using deep reinforcement learning (DRL) for bipedal robots walking in socially interactive environments. The inherent complex dynamics of bipedal robots challenge their safe maneuvering capabilities in dynamic environments. Furthermore, the intricacies of social interactions and cues such as emotions significantly compound the challenges of bipedal robot navigation. To address these coupled issues, we propose a two-stage pipeline that jointly considers the bipedal full-body dynamics and the complexities of socially aware navigation. More specifically, an emotion-integrated navigation policy is developed to balance safety, efficiency, and social courtesy by responding to human emotional cues. One key component of the policy is a novel representation of social environments using sequential LiDAR grid maps (LGMs), from which we extract latent features, implicitly including collision regions, discomfort zones determined by emotions, social interactions, and the evolving dynamics of the navigation system including robot movements and pedestrian motions. To accounts for path tracking errors and locomotion constraints during social navigation, we present an end-to-end navigation system that incorporates full-order robot dynamics during training. Finally, extensive benchmarking and sim-to-real experiments demonstrate that our method outperforms both model-based planners and DRL baselines, and generalizes effectively to real-world environments. The hardware videos and open-source code are available at \url{https://gatech-lidar.github.io/emobipednav.github.io/}.
\end{abstract}


\section{INTRODUCTION} \label{introduction}

\IEEEPARstart{T}{he} research and development of bipedal robots have garnered significant interest within the robotics community, mainly due to their human-like morphology and versatile locomotion capabilities \cite{gu2025humanoid, wu2024Infer, Muenprasitivej2024Bipedal, LiAutonomous, NarkhedeSequential, GibsonTerrain, ZhaoReactive, KulgodTemporal, WarnkeTowards, ZhaoRobust}. While a great number of studies have focused on developing stable locomotion controllers for bipedal robots, one ultimate goal is to enable bipedal robots to autonomously and courteously navigate pedestrian-populated environments within social contexts. However, achieving safe and socially aware navigation remains a formidable challenge, originating from the intricate dynamics of bipedal robots, limited maneuvering capabilities in dynamic environments, and the complexities of understanding the intentions and interactions of pedestrians.

\begin{figure}[!t]
	\centering
	\small
	\includegraphics[width=7.5cm]{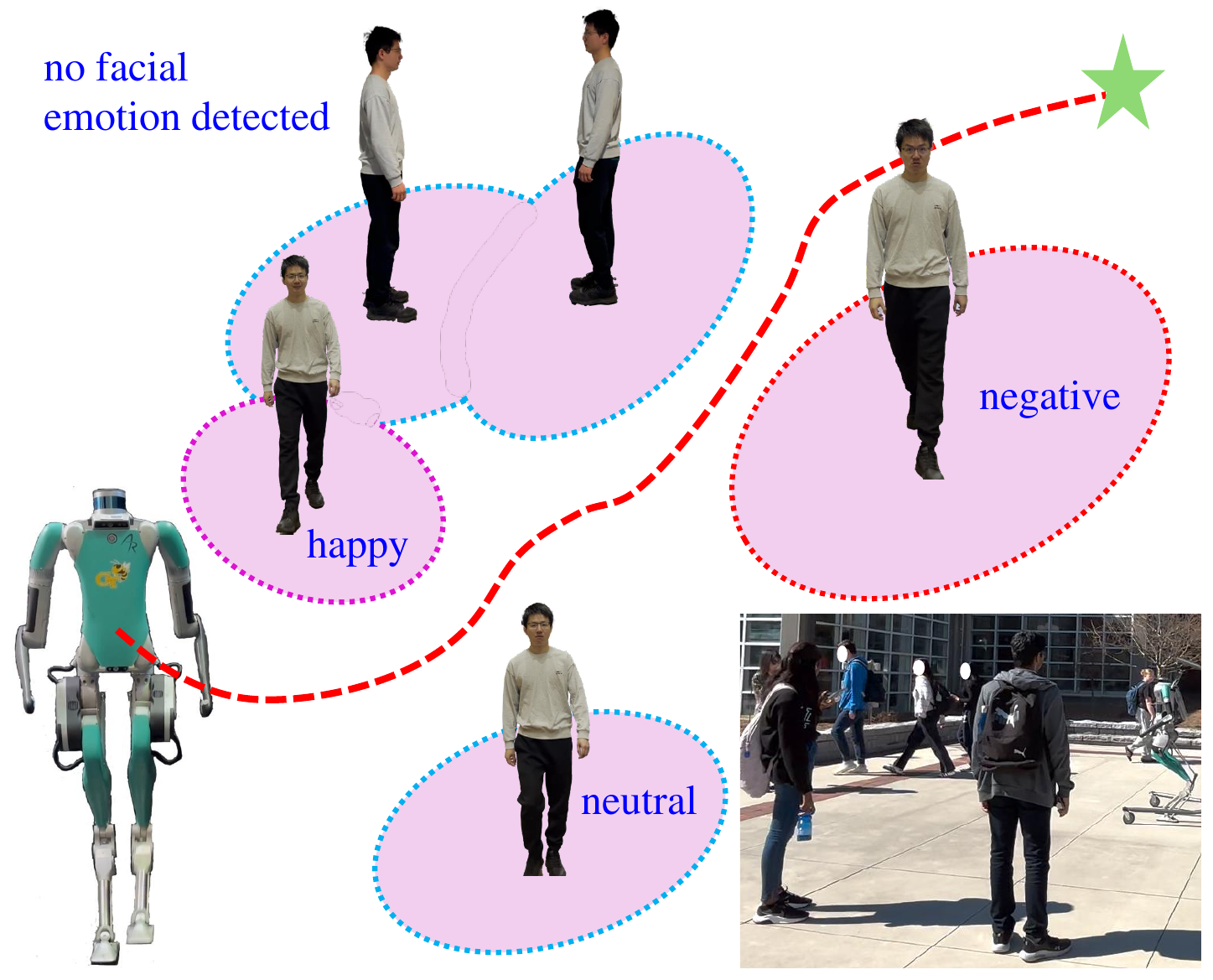}
	\caption{Emotion-aware social navigation with the bipedal robot Digit. Digit is required to maintain customized comfort distances from pedestrians with specific emotions while navigating toward a designated goal location. Note that, the comfort distances vary according to different pedestrian emotions.}
	\label{introductionFigure}
\end{figure}

Although significant efforts have been dedicated to the use of wheeled mobile robots for social navigation \cite{SingamaneniSurvey, ZhuDeep}, few studies address bipedal social navigation due to challenges in developing stable and precise locomotion controllers for their hybrid, highly nonlinear, and high-degree-of-freedom dynamics. Furthermore, complex social interactions increase the level of difficulty because they are too implicit to be accurately modeled. In addition, social cues, such as human emotions, psychologically influence the preferred personal space during interactions with other agents, including pedestrians and robots \cite{MartinezFrom, RuggieroEffect}, as illustrated in Fig. \ref{introductionFigure}. These social interactions and cues remain underexplored for bipedal robot navigation.

Cutting-edge bipedal navigation studies in crowded environments decouple navigation and locomotion, planning navigation trajectories with reduced-order models (ROMs) at first, and then tracking the trajectories with locomotion controllers based on full-order dynamics \cite{CioccaEffect, ShamsahReal, ShamsahSocially}. Such a planning architecture cannot take into account the tracking errors and constraints of low-level controllers, such as joint torque limits, for bipedal navigation. In addition, pedestrian trajectory prediction is also individually executed \cite{ShamsahSocially} because prediction models are pretrained using specific datasets only including pedestrian-pedestrian interactions, ignoring pedestrian-robot interactions. Such pretrained models might further cause deployment discrepancies in bipedal robots. Comparatively, as an end-to-end framework, deep reinforcement learning (DRL) based pipelines can account for tracking errors and constraints of low-level controllers, as well as direct interactions between pedestrians and bipedal robots, by directly training with full-body bipedal robots in pedestrian-populated environments.

In this study, we present an emotion-aware navigation framework -- EmoBipedNav -- using deep reinforcement learning (DRL) for bipedal robots walking in socially interactive environments. We represent the environment with an image-based observation of sequential LiDAR grid maps (LGMs), which include pedestrian-pedestrian and pedestrian-robot interactions, collision areas, emotion-aware discomfort zones, social interactions, and the spatio-temporal evolution of the environment. Discomfort zones can be viewed as soft collision-avoidance regions that the ego-agent should minimize intrusion into as much as possible. We employ a DRL-based navigation architecture to map these latent features directly to bipedal robot actions, specifically, the sagittal center-of-mass (CoM) speed derived from ROMs and the heading angle of Digit's torso. To address the discrepancies between ROMs and full-order dynamics, we train the social navigation policy directly in a physics-based simulator with a full-body locomotion controller, bypassing reliance on ROMs. This approach ensures that the DRL model accounts for full-body constraints and dynamics inherent to bipedal locomotion. In addition, we benchmark against state-of-the-art (SOTA) navigation pipelines to highlight the efficacy of our framework.

The main contributions of this study are as follows.

\begin{itemize}

\item \textbf{Novel DRL observation representation}. We introduce LGMs as a novel representation of highly dynamic environments integrated with pedestrian emotions. 

\item \textbf{Emotion integration in socially interactive environments}. Our framework incorporates pedestrian emotions and constructs various discomfort zones, enabling emotion-aware navigation in socially interactive environments.

\item \textbf{End-to-end navigation framework}. We propose an end-to-end navigation pipeline that directly maps latent features to bipedal robot actions and is trained with full-body bipedal robot dynamics, effectively mitigating model discrepancies induced by ROMs.

\item \textbf{Benchmarks and hardware evaluations}. We compare the performance of our method with SOTA baselines. In addition, we transfer the simulated navigation policy to the real bipedal robot Digit.
	
\end{itemize}

\section{Related Work} \label{relatedWork}
\textbf{Navigation with bipedal robots}. Numerous studies have investigated bipedal robot navigation in relatively simple environments with static and sparsely distributed obstacles \cite{LiAutonomous, TsunekawaVisual, HongReal, LiuConstraints, PengSafe}. Recent approaches have begun to address dynamic environments by incorporating moving obstacles alongside static barriers \cite{HildebrandtReal, NarkhedeSequential, ShamsahIntegrated, NarkhedeOvertaking}. However, these dynamic obstacles are sparsely distributed around the robot, e.g. only one single moving obstacle is in close proximity to the ego-agent robot. Moreover, the interactions are oversimplified, assuming slower moving obstacles and neglecting social interactions between pedestrians and robots \cite{ShamsahIntegrated}.

Research specifically addressing crowd navigation for bipedal robots remains largely underexplored, with the exception of the recent study in \cite{ShamsahReal, ShamsahSocially}, which assumes that bipedal robots and pedestrians adhere to the same social rules, that is, robots learn their navigation path from real pedestrian datasets and imitate pedestrian navigation strategies. However, this assumption is restrictive, as bipedal robots exhibit fundamentally different dynamics and face more stringent physical constraints compared to pedestrians. Furthermore, current collision avoidance strategies are typically conservative because robots risk freezing due to the stochastic and unpredictable nature of pedestrian trajectory predictions \cite{ChenDecentralized, KretzschmarSocially}.

\begin{figure*}[!t]
	\centering
	\small
	\includegraphics[width=17.0cm]{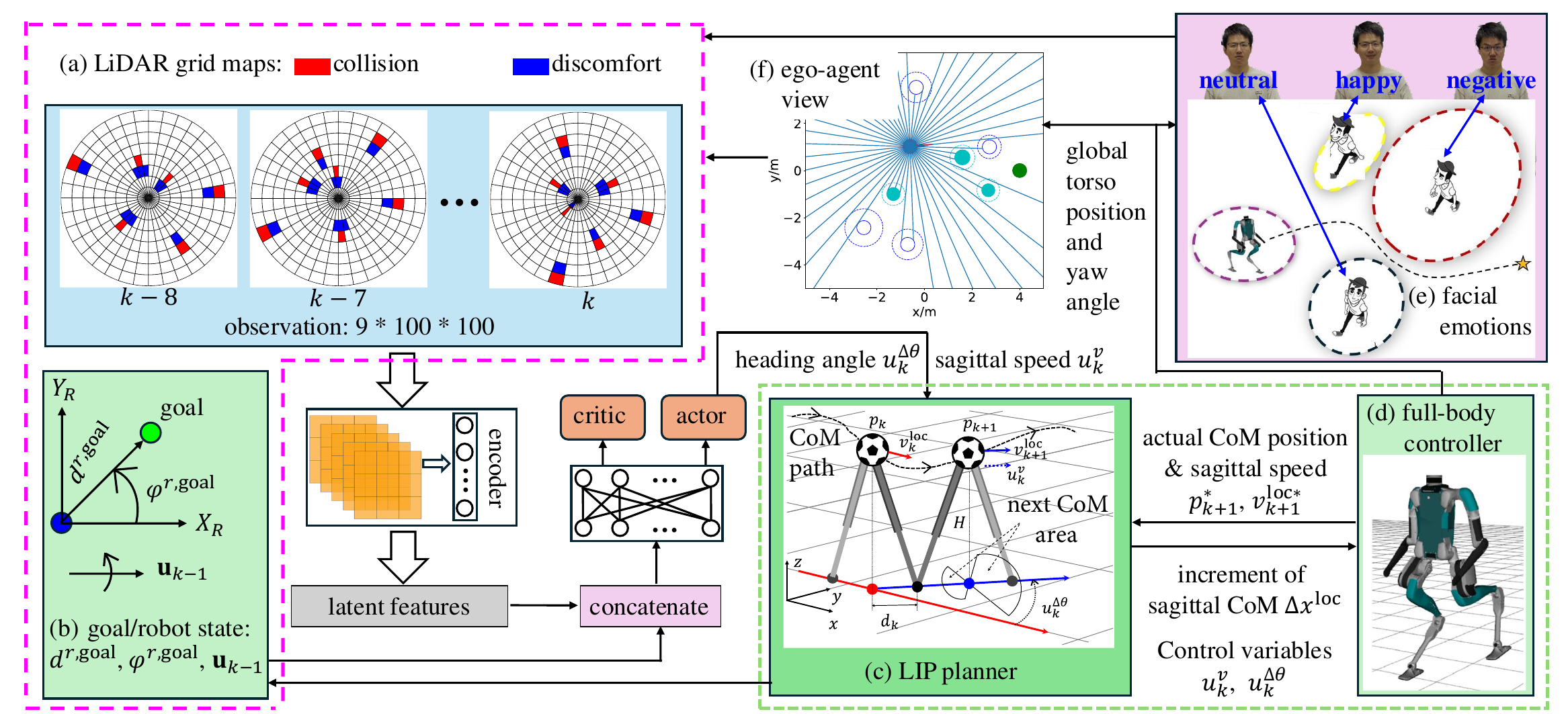}
	\caption{Overview of the proposed EmoBipedNav framework using the bipedal robot Digit. Our framework begins by obtaining estimated facial emotions (Fig. (e)) using pre-trained CNN models. Simultaneously, we transform raw LiDAR scans (Fig. (f)) into sequential pie-shape LGMs (Fig. (a)), where the red cells denote collision areas and the blue patches highlight discomfort zones associated with pedestrian emotions. These grid maps are converted into stacked pixel images which are further processed through an encoder constructed using CNNs to extract socially interactive and emotionally aware features. The resulting latent features are concatenated with the robot's last command and target position (Fig. (b)), which are fed into an actor-critic DRL structure implemented with multi-layer perceptrons (MLPs). The action output from the actor network is derived from the ROM (Fig. (c)) and practically applied to a bipedal robot with full-body dynamics and constraints (Fig. (d)). The torso position and yaw angle obtained from Digit correspond to the ego-agent state in Fig. (f). We use the Angular momentum LIP planner (ALIP) \cite{gong2021one} and a passivity full-body controller with ankle actuation \cite{ShamsahIntegrated} to track the desirable ROM trajectory.}
	\label{framework}
\end{figure*}

\textbf{Social navigation with DRL}. Although research on social navigation with bipedal robots is limited, there is a substantial body of work on social navigation for wheeled mobile robots \cite{SingamaneniSurvey}. In particular, end-to-end frameworks that leverage DRL have been widely used to directly map environmental states to robot actions \cite{ZhuDeep}. A prominent avenue is the collision avoidance approach with deep reinforcement learning (CADRL) \cite{ChenDecentralized}, which formulates the problem as a Markov Decision Process (MDP) and assumes fully observable environments, including pedestrian attributes such as size, position, velocity, and preferred speed. Moreover, CADRL only pairs the ego-agent with each pedestrian individual, neglecting pedestrian interactions. Subsequent studies, such as the social attention RL (SARL) \cite{ChenCrowd} and the relational graph learning (RGL) \cite{ChenRelational}, focus on modeling implicit pedestrian interactions to plan socially acceptable motions. However, these SOTA models rely heavily on fully observable environments which require complicated perception and prediction techniques to obtain in real scenarios. In addition, they assume that the kinematics of mobile robots are less constrained and similar to those of pedestrians, which is not applicable to bipedal robots due to their restrictive locomotion capabilities.

While MDPs are designed for decision making in fully known environments, Partially Observable MDPs (POMDPs) are suitable for decision making in socially interactive environments, which involves partially observable pedestrian interactions and future trajectories \cite{KaelblingPlanning}. SOTA approaches that map raw or preprocessed sensor data to robot actions can be formulated as POMDPs \cite{FanDistributed, JinMapless, ZhuAutonomous, ZhuLearn, XieLearning, LianTransferability}. An intuitive representation is the sequential LiDAR scans, which are stacked and encoded with convolutional neural networks (CNNs) \cite{FanDistributed, JinMapless} or directly processed with long-short-term memory (LSTM) \cite{ZhuLearn}. Combining sequential LiDAR scans with fully known states (i.e., pedestrian position and speed), the DRL velocity obstacle (DRL--VO) method converts these observations and states into an image for feature extraction \cite{XieLearning}. In practice, they are dependent on complex estimation techniques to obtain these states \cite{XieLearning}. Moreover, their network architectures are susceptible to divergence, plateauing, and collapse when applied to robots with complicated dynamics and strict constraints \cite{JinMapless, ZhuLearn, XieLearning}. Compared to others, our approach is more straightforward, with only sequential LiDAR scans converted into grid maps. In addition, the neural networks of our feature extractor followed by the soft actor-critic (SAC) DRL algorithm \cite{HaarnojaSoft} yield a minimal risk of convergence failures for DRL training. 

\textbf{Emotion-aware navigation}. Most of the studies discussed above simplify pedestrians as circles, overlooking their social cues such as emotions. Pedestrian emotions implicitly convey information, such as the preferred comfort distance that people prefer to maintain from others \cite{MartinezFrom}. Psychological experiments \cite{RuggieroEffect} have identified comfort distances associated with various emotions. With these insights, social robots can adaptively maintain a customized comfort distance from pedestrians with specific emotions \cite{NarayananGait, NarayananEmotion}. However, these emotion-aware robot navigation approaches focus primarily on the emotions of individuals, neglecting the complex interactions that occur among pedestrians with various emotional states, which will be addressed in our study.

\vspace{-2mm}

\section{APPROACH} \label{approach}
In this section, we first introduce the reduced-order model (ROM) of bipedal locomotion and the Partially Observable Markov Decision Process (POMDP) of the social navigation problem. Subsequently, we present our DRL-based approach, detailing the novel observation space, the DRL network architecture, and the design of the integrated reward functions. The overall framework is illustrated in Fig. \ref{framework}.

\vspace{-1mm}

\subsection{Preliminaries}
The dynamics of bipedal robots is commonly approximated using ROMs, such as the linear inverted pendulum (LIP) model \cite{gong2021one, KajitaLinear}, as depicted in Fig. \ref{framework}-(c). The discrete state transition equation for the LIP model at the $k^{\rm th}$ walking step in the local sagittal frame is given as follows:
\begin{equation} 
\begin{split}
d_{k} & = \frac{\upsilon _k^{\text{loc}} \cosh \left(\omega T\right) - u_{k}^{\upsilon}} {\omega \sinh \left(\omega T\right)},\\
\Delta x^{\text{loc}}(d_{k}) & = \upsilon _k^{\text{loc}} \frac{\sinh \left(\omega T\right)}{\omega} + d_{k} \left(1 - \cosh \left(\omega T\right)\right), 
\end{split}
\label{EqLipModel}
\end{equation}
where the control input $u_{k}^{\upsilon}$ represents the desired sagittal velocity at the next foot placement switch instant, the output state $\Delta x^{\text{loc}}(d_{k})$ is the local sagittal CoM position increment between two consecutive walking steps, $\upsilon _{k}^{\text{loc}}$ denotes the sagittal velocity in the local coordinate at the $k^{\rm th}$ walking step switch instant, $d_{k}$ is an auxiliary variable that specifies the sagittal foot distance relative to CoM, $T$ is the constant footstep time, the term $\omega=\sqrt{g/H}$ is derived from the gravitational constant $g$ and the desired CoM height $H$. The ROM in (\ref{EqLipModel}) is equivalent to the formulation of the angular momentum LIP planner (ALIP) model in \cite{gong2021one}. However, since our DRL network outputs the sagittal speed, our formulation uses the current sagittal velocity of CoM, $\upsilon _k^{\text{loc}}$, instead of the angular momentum about the contact point used in ALIP.

Since $u_{k}^{\upsilon}$ influences only the translational motion, we introduce a heading angle $u_{k}^{\Delta\theta}$ to steer the CoM orientation in the horizontal plane, as illustrated in Fig.~\ref{framework}-(c). Combining the translational and rotational motions of the CoM, we define the action of our navigation policy as:
\begin{equation}
    \mathbf{u}_{k} = \{u_{k}^{\upsilon}, u_{k}^{\Delta\theta}\}, ~0 \leq u_{k}^{\upsilon} \leq \bar{u}^{\upsilon}, ~\mid u_{k}^{\Delta\theta} \mid \leq \bar{u}^{\Delta\theta},
    \label{EqAction}
\end{equation}
where $\bar{u}^{\upsilon}$ and $\bar{u}^{\Delta\theta}$ represent the maximum allowable sagittal speed and heading angle, respectively. 

Given the control $\mathbf{u}_{k}$, the CoM displacement $\Delta x^{\text{loc}}$, and the LIP model, the CoM position is propagated from $p_k$ to $p_{k+1}$, and the sagittal speed changes to $\upsilon _{k+1}^{\text{loc}}$ from $\upsilon _{k}^{\text{loc}}$. However, there are discrepancies between the actual CoM position $p_{k+1}^{*}$ and $p_{k+1}$, and between the actual sagittal speed $\upsilon _{k+1}^{\text{loc*}}$ and $\upsilon _{k+1}^{\text{loc}}$. More specifically, the full-body controller generates joint torques to track the desired CoM displacement $\Delta x^{\text{loc}}$ and the control $\mathbf{u}_{k}$, as illustrated in Fig. \ref{framework}-(d). However, discrepancies between the LIP model and the full-body dynamics can cause the robot’s states to deviate from the desired ones, which may lead to collisions and intrusions into discomfort areas of pedestrians, as demonstrated in Section \ref{experiments}. Furthermore, due to the limitations of Digit's hardware and locomotion controllers, these motion constraints are significantly more stringent than those of pedestrians, making Digit less reactive to pedestrians with high walking speed (e.g., $1$ m/s), thus posing additional challenges for Digit navigating through crowded pedestrian environments. In Section \ref{experiments}, we show the benchmark results to demonstrate that these constraints significantly degrade the navigation performance of the SOTA approaches.

\textbf{Remark 1: }\textit{In our previous work \cite{ShamsahReal, ShamsahSocially}, the ROM in conjunction with a model predictive controller is used to plan feasible CoM trajectories for Digit. However, in this work, the ROM state (i.e., the heading angle and sagittal speed in Eq. (\ref{EqLipModel})) is used to define the DRL actions. The input of the proposed network comes from the full-body simulation of Digit (Fig. \ref{framework}(d)), eliminating any discrepancies between the ROM and full-body dynamics, which exist in our previous work. Furthermore, we utilize the ROM for benchmarks, including both model-based motion planners and DRL-based baselines, effectively demonstrating the practical impact of robot dynamics discrepancies.}

\subsection{Problem Formulation}

We formulate the social navigation problem as a POMDP because the state information of social environments is not fully observable, such as implicit social interactions and predicted pedestrian trajectories. More specifically, we represent social environments with sequential LiDAR grid maps (LGMs). Despite only including partial information, LGMs allow the extraction of implicit features such as collision zones, pedestrian-pedestrian and pedestrian-robot interactions, emotion-related discomfort areas, and the spatio-temporal dynamic evolution of the navigation system.

In addition to LGMs which only contain partial information, we have two other fully observable states. For one thing, the goal position state is responsible for the target-directed navigation task. In addition, since pedestrians interact with the robot and the robot's actions influence pedestrian decisions, we also include the robot's actions as part of the states.

Given the observation of LGMs and the fully observable states, our objective is to optimize a navigation policy, straightforwardly projecting the observation and states into the robot actions. The problem is formulated as a POMDP defined by a 6-tuple $\left(S, O, A, \Gamma, R, \gamma\right)$, where $S$ is the fully observable state space including the goal position and the robot's actions, $O$ represents the observation space consisting of sequential LGMs, $A$ stands for the action space, $\Gamma$ denotes the state transition model fully propagated with the bipedal robot Digit as shown in Fig. \ref{framework}-(d), $R$ is the reward function with an instantaneous scalar reward $r_k$ at the $k^{\rm th}$ walking step and $\gamma$ is the discount factor. To derive the optimal navigation policy $\pi(a \in A | s \in S, o \in O)$, our goal is to maximize the expected sum of discounted rewards over an infinite time horizon, as follows:

\begin{equation}
    V = \mathbb{E} _{\pi} \left[ \sum_{k=0}^{\infty}\gamma ^k r_k \right].
    \label{EqDiscountReturn}
\end{equation}

We will leverage a CNN-based encoder followed by a soft actor-critic (SAC) DRL algorithm \cite{HaarnojaSoft} to iteratively optimize the navigation policy $\pi$.

\subsection{Observation and State Spaces}

\begin{figure}[!t]
	\centering
	\small
	\includegraphics[width=8.5cm]{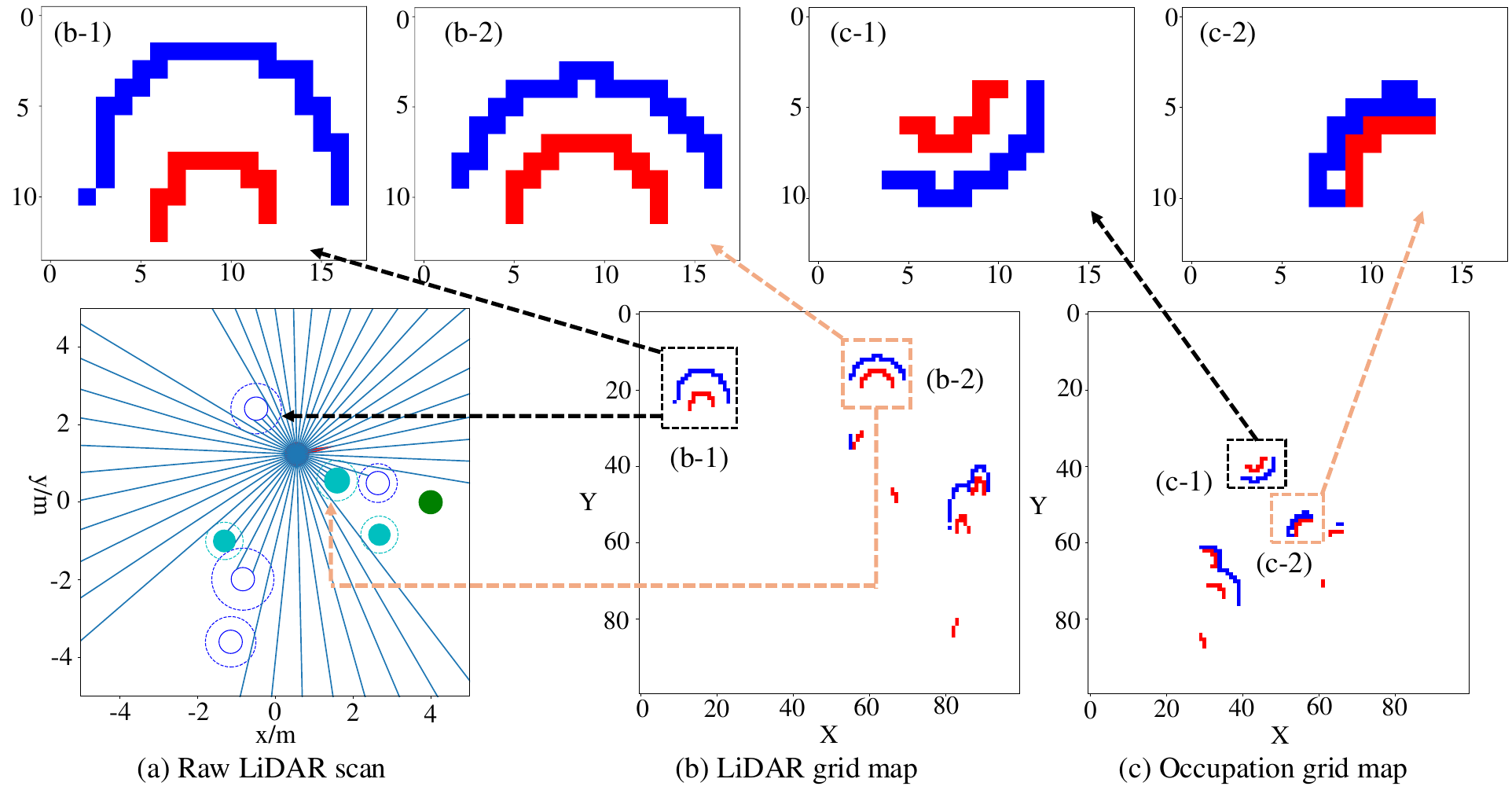}
	\caption{LiDAR scan and its corresponding grid maps. (a) illustrates the raw LiDAR scan. The red solid circle with an arrow represents the ego-agent and its orientation in the world frame, and the green solid one is the goal. Moving pedestrians are blue hollow circles, and static obstacles are depicted as solid cyan circles. The outer dashed circles illustrate the discomfort margins. (b) and (c) are the corresponding LiDAR grid map and occupation grid map, respectively. X and Y axes in (b) represent rotational and radial indices, while X and Y in (c) are horizontal and vertical pixel positions, respectively. (b-1) and (b-2) shown in (b) are the margins corresponding to two close objects surrounding the ego-agent illustrated in (a). (c-1) and (c-2) correspond to the same objects as (b-1) and (b-2), respectively. Their zoomed-in views are shown on the top row.}
	\label{GridMap}
\end{figure}

We propose a novel representation for the observation space, referred to as pie-shape LGMs (Fig. \ref{framework}-(a)), to comprehensively encapsulate the features of all entities around the ego-agent, including static obstacles having variable sizes and dynamic pedestrians with different emotions. Sequential pie-shape LGMs are built on the basis of raw LiDAR scans. More specifically, a 2-D LiDAR scan with $K$ beams is emitted from the ego-agent to detect collision and discomfort margins, as illustrated in Fig. \ref{framework}-(f) and Fig. \ref{GridMap}-(a).

\textit{Remark 2. During training, we assume pedestrian emotions are pre-known and randomly assign them to facilitate sim-to-real transfer. These emotions are used to define corresponding discomfort zones, which contribute to the construction of the LiDAR grid maps. In real-world implementations, pedestrian emotions are estimated from facial cues using a pre-trained CNN model \footnote{\label{emotionNet}https://github.com/susantabiswas/realtime-facial-emotion-analyzer} consisting of six convolutional layers and four fully connected layers.}

The detected obstacles are projected onto a pie-shape grid map (shown in Fig. \ref{framework}-(a)) with a maximum detection radius of $L$. This map is divided into $M$ radial segments, forming $M$ concentric rings. Each ring is further segmented angularly into $M$ equal sections. The resolution of each grid increases as the ring gets closer to the center, ensuring that more details around the ego-agent are detected. Note that the pie-shape grid map is inherently circular and cannot be directly used for feature extraction with CNNs, which require a rectangular image input. To address this, the map is first flattened and restructured into a LiDAR grid map, as shown in Fig. \ref{GridMap}-(b).

Let $g\left(i, j\right)$ denote the grid cell in the $i^{\rm th}$ ring and the $j^{\rm th}$ angular segment. Each grid cell is intersected by $K / M$ LiDAR beams passing through it. Consequently, we define the resolution of each grid cell as the number of beams per unit area, expressed as:
\begin{equation}
    g_r = \frac{KM^2}{\pi L^2 \left(2i - 1\right)},
    \label{EqGridResolution}
\end{equation}
which ensures that more detailed information, such as the collision cells and discomfort patches shown in Fig. \ref{framework}-(a), is captured for objects closer to the robot, allowing for more accurate local navigation around the ego-agent. Note that the pie-shape grid map is inherently circular and cannot be directly used for feature extraction with CNNs, which  require a rectangular image input. To address this, the map is first flattened and restructured into a LiDAR grid map, as shown in Fig. \ref{GridMap}-(b).

Alternatively, occupancy grid maps (OGMs) \footnote{http://wiki.ros.org/costmap$_-$2d}, as shown in Fig. \ref{GridMap}-(c), are commonly used for local collision avoidance \cite{ElfesUsing}. With the same map size $M^2$ and detection radius $L$, the resolution of each grid is calculated evenly as $(M/2L)^2$. In contrast, the resolution of our LGM is calculated on average as $(M/\sqrt{\pi}L)^2$, demonstrating that our LGM representation has a higher resolution on average. More specifically, the LGM has a higher resolution for grids closer to the map origin and a lower resolution for grids farther away. For example, when $M=100$, $L=6.0$ m, and $K=1800$, for $i=1$, the LGM has 100 grids while the OGM only has 4. When $i > 13$, the density of the OGM grid surpasses that of the LGM. As a result, the LGM can represent more details within the $13 \cdot L / M = 0.78$ m range of the ego-agent, as illustrated in the top figures of Fig. \ref{GridMap}. Given the same obstacles close to the ego-agent, their representation in Fig. \ref{GridMap}-(b) (LGM) contains more pixels compared to Fig. \ref{GridMap}-(c) (OGM), indicating that LGM can capture more details when obstacles are near, which is beneficial for collision avoidance and alleviating discomfort, as demonstrated in Section \ref{experiments}.

Psychological experiments \cite{RuggieroEffect} indicate that pedestrians feel comfortable maintaining a social distance from others, and this distance varies depending on their emotional state. For example, people tend to increase distance when encountering angry or sad individuals compared to those who appear happy. Motivated by this, we aim to integrate emotion-related distance as an additional observation to enable emotion-aware navigation. In particular, we define a discomfort margin (blue grids in Fig. \ref{framework}-(a) and Fig. \ref{GridMap}-(b)) that expands according to the collision area (red cells in Fig. \ref{framework}-(a) and Fig. \ref{GridMap}-(b)). The size of this discomfort zone is adjusted according to the emotional state of the pedestrian. We consider three emotional states: \texttt{happy}, \texttt{neutral}, and \texttt{negative} including angry, sad, fearful, and surprised. According to psychological research \cite{RuggieroEffect}, happy emotions encourage approach behaviors and therefore shorter distances, whereas negative emotions promote avoidant behaviors and longer distances. Based on the studies in \cite{RuggieroEffect, NarayananEmotion}, we assign the following distances from the robot margin to the pedestrian edge: $0.2$ m for \texttt{happy}, $0.35$ m for \texttt{neutral}, and $0.5$ m for \texttt{negative} emotions.

The observation $\mathbf{o}_k$ at the current walking step is the LGM capturing only the static immediate environment in the spatial dimension, which might encourage conservative motions to avoid collisions as quantitatively analyzed in Section \ref{experiments}. To capture dynamic information in the temporal dimension, such as pedestrian-pedestrian and pedestrian-robot interactions, and the evolving dynamics of the navigation system, we incorporate previous LGMs from time step $k-N+1$ to $k-1$ as illustrated in Fig. \ref{framework}-(a). Thus, the observation in both the spatial and temporal dimensions is represented as $\mathbf{o}_k^{\rm{sp-tem}} = \{\mathbf{o}_{k-N+1}, ..., \mathbf{o}_{k-1}, \mathbf{o}_k\}$. From spatio-temporal LGMs, we can extract latent features, including collision areas, complex pedestrian-pedestrian and pedestrian-robot interactions, emotion-related discomfort zones, and the dynamic evolution of the navigation system.

In addition to the observation only including partial information, we define another two fully observable states as illustrated in Fig. \ref{framework}-(b): the position state of the goal $\mathbf{s}_k^{\rm{goal}}=\{d^{r,{\rm goal}}, \varphi^{r,{\rm goal}} \}$, where $d^{r,{\rm goal}}$ represents the distance between the goal and the robot and $\varphi ^{r, {\rm goal}}$ indicates the orientation of the goal in the robot frame; the last robot command $\mathbf{u}_{k-1}$. These two states are used for the goal-reaching task and the pedestrian-robot interaction, respectively.

\subsection{Network Structure}
The encoder, as shown in Fig. \ref{framework}, consists of CNNs followed by multi-layer perceptrons (MLPs) to fully extract the latent features $\mathbf{f}_k^o$ from the sequential LGMs $\mathbf{o}_k^{\rm{sp-tem}}$. These features inherently capture pedestrian-pedestrian and pedestrian-robot interactions, obstacle positions, emotion-related discomfort zones, and the dynamic evolution of the navigation system. Unlike SOTA approaches \cite{ChenCrowd, ChenRelational} that rely solely on MLPs to process fully observable states, our encoder is capable of extracting more detailed and implicit information from observation LGMs. Additionally, the proposed image-based observation integrates emotion-related discomfort zones, addressing a challenge faced by cutting-edge methods using raw sensor data \cite{FanDistributed, JinMapless}.

Given these latent features $\mathbf{f}_k^o$, we concatenate them with the goal state $\mathbf{s}_k^{\rm{goal}}$ and the action state $\mathbf{u}_{k-1}$. The combined inputs are fed into the SAC DRL framework \cite{HaarnojaSoft} to optimize the navigation policy $\pi\left(\mathbf{u}_{k}|\mathbf{o}_k^{\rm{sp-tem}}, \mathbf{s}_k^{\rm{goal}}, \mathbf{u}_{k-1}\right)$. The actor and value modules, as shown in Fig. \ref{framework}, are made up of MLPs that take the same inputs $\{\mathbf{f}_k^o, \mathbf{s}_k^{\rm{goal}}, \mathbf{u}_{k-1}\}$. In summary, the network structure can be expressed as follows:
\begin{subequations}
\label{EqNetwork}
    \begin{align}
        \mathbf{f}_k^o & = f_\vartheta(\mathbf{o}_k^{\rm{sp-tem}}),  \label{Encoder} \\
        v_k & = f_\phi(\mathbf{f}_k^o, \mathbf{s}_k^{\rm{goal}}, \mathbf{u}_{k-1}), \label{Critic}\\
        \mathbf{u}_{k}& \sim f_\psi(\mathbf{f}_k^o, \mathbf{s}_k^{\rm{goal}}, \mathbf{u}_{k-1}), \label{Actor}
    \end{align}
\end{subequations}
where the network parameters $\vartheta, \phi, \psi$ are updated using the SAC algorithm \cite{HaarnojaSoft} and the exploration samples collected from the bipedal robot Digit, $v_k$ represents the deterministic value function predicted by the critic network, and $\mathbf{u}_{k}$ is the action sampled from a Gaussian distribution with its mean and standard deviation parameters output from the actor network. Unlike the original SAC framework, our approach introduces a shared encoder network (\ref{Encoder}) that is used by both the critic network (\ref{Critic}) and the actor network (\ref{Actor}). This shared encoder is updated exclusively with the critic network, as it is primarily responsible for evaluating robot behaviors as well as the corresponding observation and states. The encoder is then copied to the actor network without gradient computations, ensuring that the feature extraction process remains consistent across both networks.
 
\subsection{Reward Integration}

The objectives of our navigation task are threefold: avoidance of collisions, reaching goals, and awareness of emotions. Accordingly, we define the reward function as follows:
\begin{equation}
    r_k = r_k^{\rm{col}} + r_k^{\rm{goal}} + r_k^{\rm{emo}},
    \label{EqRewardFunctuin}
\end{equation}
where $r_k^{\rm{col}}$ penalizes the collisions with all surrounding objects, $r_k^{\rm{goal}}$ rewards progress toward reaching the goal, and $r_k^{\rm{emo}}$ denotes an emotion-related penalty. Specifically, these three rewards are defined as follows.

\begin{figure}[!t]
	\centering
	\small
	\includegraphics[width=7.0cm]{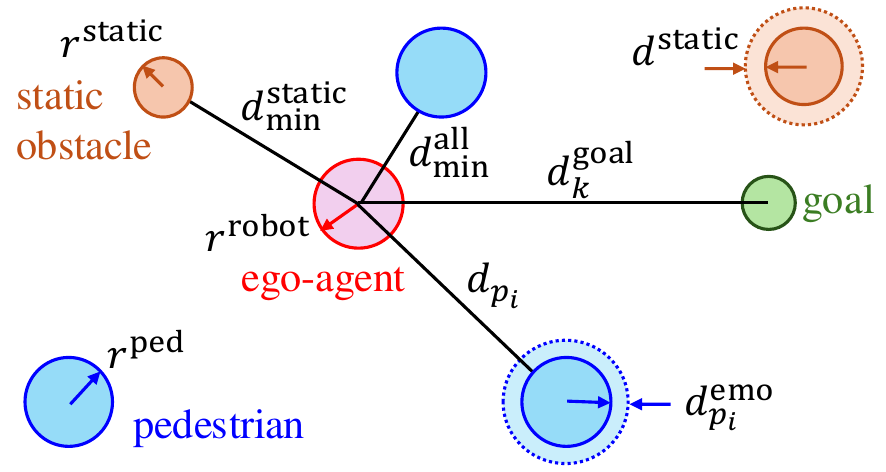}
	\caption{Geometry representations of reward-related notations.}
	\label{rewardDis}
\end{figure}

\noindent 1) Collision avoidance reward:
\begin{equation}
r_k^{\rm{col}} =
\begin{cases}
-0.6 & \text{if} ~ d^{\rm{all}}_{\min} \leq r^{\rm{robot}},\\
-0.1 & \text{else if} ~ d^{\rm{static}}_{\min} < d^{\rm{static}} + r^{\rm{robot}}, \\
0.0 &  \text{else}, \\
\end{cases}   
\label{EqRewardCollision}   
\end{equation}
where $d^{\rm all}_{\min}$ represents the minimum distance from the ego-agent to the margins of all surrounding objects as shown in Fig. \ref{rewardDis}, $d^{\rm static}_{\min}$ denotes the minimum distance from the ego-agent to all static obstacles, $d^{\rm{static}}$ stands for a safety distance expanded from the collision margin of the obstacle, and $r^{\rm{robot}}$ is the ego-agent radius.

\noindent 2) Goal reaching reward:
\begin{equation}
r_k^{\rm{goal}} =
\begin{cases}
0.5 & \text{if} ~ d_k^{\rm{goal}} \leq d^{\rm{goal}},\\
0.3 \cdot (d_{k-1}^{\rm{goal}} - d_k^{\rm{goal}}) & \text{else}, \\
\end{cases}   
\label{EqRewardGoal}   
\end{equation}
where $d_{k}^{\rm{goal}}$ stands for the distance from the ego-agent to the goal at the $k^{\rm th}$ walking step, and $d^{\rm{goal}}$ represents the goal-reaching threshold. In particular, the second term encourages the robot to move toward the goal, thus alleviating the sparsity issue of the reward function and promoting efficient training.

\noindent 3) Emotion-aware reward:
\begin{equation}
r_k^{\rm{emo}} =
\begin{cases}
-0.1 & \text{if} ~ \mathcal{C},  \\
0.0 & \text{else},\\
\end{cases}   
\label{EqRewardEmotion}   
\end{equation}
where $\mathcal{C}$ is a distance condition with respect to pedestrian emotions. Specifically, the discomfort distance of the $i^{\rm th}$ pedestrian is defined as $d_{p_i}^{\rm{emo}} \in \{0.2, 0.35, 0.5\}$m, depending on the corresponding emotion of the pedestrian $\{\texttt{happy}, ~\texttt{neutral}, ~\texttt{negative}\}$. $r^{\rm{ped}}$ is a constant pedestrian radius. Given the distance $d_{p_i}$ from the margin of the $i^{\rm th}$ pedestrian to the ego-agent, the penalty is $-0.1$ if $\exists ~i$, such that $d_{p_i} < d_{p_i}^{\rm{emo}} + r^{\rm{robot}}$. Note that, to avoid learning a conservative navigation policy, we only count once if the ego-agent enters the discomfort zones of several objects, including static obstacles and dynamic pedestrians, at the same walking step. That means we add $-0.1$ only once in Eq. (\ref{EqRewardFunctuin}) although it appears in both Eq. (\ref{EqRewardCollision}) and Eq. (\ref{EqRewardEmotion}).

\textit{Remark 3. Identifying the optimal reward configuration is challenging due to the highly coupled relationships among different reward terms. For example, when larger penalty factors were applied for entering discomfort zones (e.g., -0.6), the number of discomfort events decreased; however, this came at the cost of increased navigation time, as the agent adopted more conservative behaviors. Due to space limitations, we have chosen not to include the detailed results of these parameter variations. Instead, we empirically fine-tuned the reward function and ultimately selected a feasible and sub-optimal combination that balances safety and efficiency.}

Our algorithm is summarized in \textbf{Algorithm} \ref{alg:emobipednav}.

\begin{algorithm}
\caption{EmoBipedNav Algorithm}
\label{alg:emobipednav}
\begin{algorithmic}[1]
\STATE \textit{Initialize}: SAC replay buffer $\mathcal{D}$, $k \leftarrow 1$
\STATE \textit{Reset}: pedestrian positions and emotions, positions and sizes of static obstacles, robot position, and $\mathbf{u}_{k-1} \leftarrow \mathbf{0}$
\STATE \textit{Observe}: $\mathbf{o}_k^{\rm{sp-tem}}$, $\mathbf{s}_k^{\rm{goal}}$, and $\mathbf{u}_{k-1}$
\WHILE{$k \leq \mathcal{K}$}
    \STATE Sample action $\mathbf{u}_{k}$ using Eq. (\ref{Encoder}) and (\ref{Actor})
    \STATE Compute the desired foot placement with ALIP \cite{gong2021one} given $u_{k}^{\upsilon} \in \mathbf{u}_{k}$
    \STATE Track the full-body trajectory planned based on the foot placement using a passivity-based controller \cite{SadeghianPassivity}
    \STATE Rotate hip yaw-joint to track heading angle $u_{k}^{\Delta\theta} \in \mathbf{u}_{k}$
    \STATE Joint torque control for Digit
    \STATE $k \leftarrow k+1$
    \STATE \textit{Observe}
    \STATE Calculate reward with Eq. (\ref{EqRewardFunctuin})-(\ref{EqRewardEmotion})
    \IF{goal reached \OR collision \OR timeout}
        \STATE \textit{Reset}
    \ENDIF
    \STATE Add observation, reward, and reset information to $\mathcal{D}$
    \STATE Update networks Eq. (5) using samples from $\mathcal{D}$ and SAC Algorithm
\ENDWHILE
\end{algorithmic}
\end{algorithm}

\section{Experiments} \label{experiments}

\renewcommand\arraystretch{1.0}
\begin{table}[!t]
	\small
	\centering
	\caption{Parameter (Param) Set-up.}
	\begin{tabular}{c c | c c | c c} 
		\hline
		Param     &  Value    & Param              &  Value      & Param    &  Value    \\  
		\hline 
		$T$       &  $0.4$ s  & $H$                & $1.02$ m    &  $g$     & $9.81$ m/s$^2$\\
            $K$       &  1800     & $M$                & 100         &  $L$     & $6.0$ m\\
            $\gamma$  & 0.99      & $d^{\rm{static}}$  & $0.2$ m     &  $\bar{u}^{\upsilon}$ & $0.4$ m/s \\
            $N$       & 9         & $d^{\rm{goal}}$    & $0.1$ m     &  $\bar{u}^{\Delta\theta}$ & $0.2$ rad  \\
            $r^{\rm{ped}}$  & $0.3$ m      & $r^{\rm{static}}$  & $[0.2, 0.4]$ m & $r^{\rm{robot}}$ & $0.3$ m \\
		\hline
	\end{tabular}
	\label{tableParameter}
\end{table}

We first detail the parameter specifications and the network architecture, followed by the environmental settings. Subsequently, we experimentally evaluate the tracking errors introduced by the use of ROMs. Next, we benchmark our proposed navigation approach against SOTA methods that do not account for emotion-related observations and rewards. In addition, we implement another training with emotion-related observations and reward functions. Further investigations on time-varying emotions are carried out. Finally, we demonstrate the deployment of our navigation policy on the bipedal robot Digit.

\subsection{Parameter Specifications and Network Details}
TABLE \ref{tableParameter} presents the parameter specifications. Our DRL network architecture consists of 3 modules: encoder, actor, and critic. The encoder comprises 4 layers of CNNs, each with 32 kernels of size 3. The stride of the first layer is 2, while the remaining layers have a stride of 1. After the CNN layers, a fully connected layer extracts 50 features from the sequential LiDAR maps. Both the actor and critic networks use similar MLP architectures, each with two hidden layers of size 1024. Unlike critic networks, the output of the network has only one value, while the network of actors produces 4 variables, corresponding to 2 means and 2 $\log$ standard deviations for the action $\mathbf{u}_{k}$. The learning rates for both the actor and critic networks are set to 0.001. In particular, the encoder network is updated jointly with the critic network but is detached from the actor network, avoiding overfitting to specific actions that the actor network frequently explores.

\subsection{Environment Settings}
The simulation environment includes both static obstacles and dynamically moving pedestrians interacting with a full-body bipedal robot.

\textbf{Static obstacles}. Static obstacles are randomly placed within a rectangular area, with the $x$-coordinate ranging from $-3$ to $3$ m and the $y$-coordinate from $-1.5$ to $1.5$ m. Each obstacle has a radius ranging from $0.2$ to $0.4$ m.

\textbf{Pedestrians}. The initial position of each pedestrian is distributed along the perimeter of a circle with a radius randomized from $3.5$ to $4.5$ m. The corresponding goal of each pedestrian is located on the other side of the circle. Pedestrians move between these two points at a preferred speed of $1$ m/s. 

Pedestrian motion is generated using the Optimal Reciprocal Collision Avoidance (ORCA) method \cite{BergReciprocal}, a multi-agent motion planner commonly used to simulate interactive motions of all agents \cite{ChenDecentralized, ChenCrowd, ChenRelational, JinMapless, XieLearning}. In addition to pedestrian-pedestrian interactions, we also enable pedestrian-robot interactions when the pedestrian's speed exceeds the robot's speed. Otherwise, the robot remains invisible to pedestrians, and only the robot actively avoids collisions with pedestrians. Such scenarios where pedestrians are not always trying to avoid the ego-agent are designed to independently demonstrate the active collision avoidance capability of our navigation policy. 

\textbf{Robot}. The robot navigates from position $(-4, 0)$ to position $(4, 0)$ m, interacting with pedestrians and actively avoiding static obstacles. We simulate the bipedal robot Digit in MuJoCo \cite{TodorovMujoco}. The navigation policy in the high-level motion planner only generates the sagittal speed and heading angle derived from the ROM.

For tracking sagittal speed, we first use the ALIP planner \cite{gong2021one} to generate foot placements. To address gaps in robot dynamics, we incorporate a passivity-based controller \cite{SadeghianPassivity}, which effectively preserves the natural dynamics of the system while ensuring asymptotic tracking performance. In our previous work \cite{ShamsahSocially, ShamsahIntegrated}, the addition of toe-joint actuation alongside the whole-body passivity controller demonstrated promising results in minimizing the mismatch between the LIP model trajectory and the full-order trajectory of the Digit bipedal robot. Despite mitigating these dynamics gaps, the tracking error in the sagittal velocity persists, especially in the peaks and valleys \cite{ShamsahSocially}. To track the heading angle, we use the hip yaw-actuator to align the Digit's torso to the desired CoM direction. However, as demonstrated in our previous study \cite{ShamsahSocially}, the tracking error in the torso orientation of Digit remains detrimental, especially when there are frequent changes in the desired torso orientation. 

These inaccurate trackings negatively impact social navigation performance when deploying the motion planner optimized with ROMs on the full-body bipedal robot Digit, a practical issue we will further investigate in Section \ref{experiments}-C and \ref{experiments}-D. To address these tracking errors, we directly train our navigation policy with the full-body bipedal robot, allowing it to account for tracking errors in both sagittal velocity and heading angle.

\subsection{Model Discrepancies}

We describe the model discrepancies between the ROM and full-body bipedal dynamics in Section \ref{approach}-A. To clearly and intuitively illustrate these model gaps, we conduct simulation experiments that evaluate the tracking performance.

\begin{figure}
    \vspace{0.2cm}
    \centering
    \includegraphics[width=7.5cm]{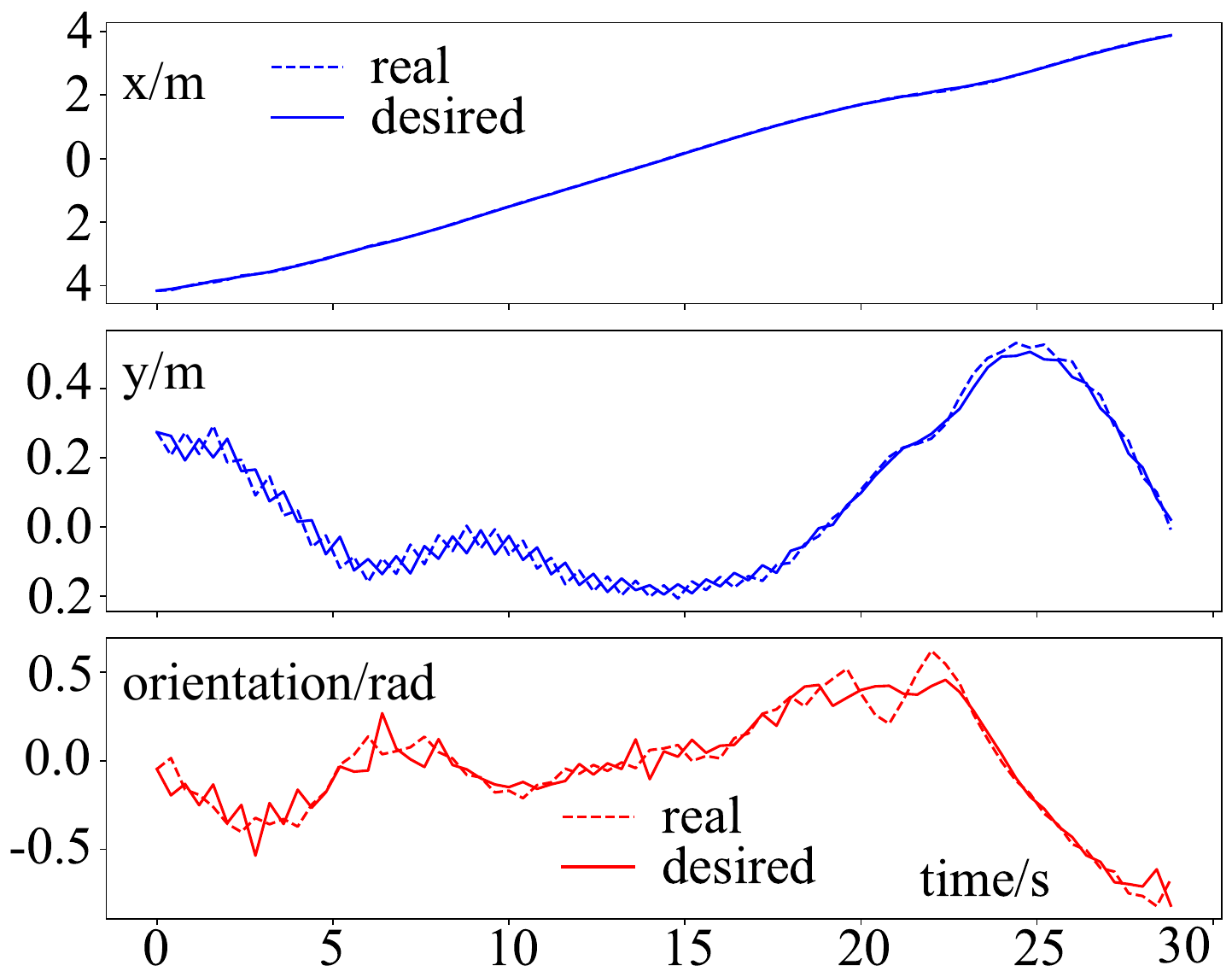}
    \caption{Tracking performance. The top figure shows the $x$ tracking, the middle plots the $y$ tracking, and the bottom illustrates the heading angle following in the world frame.}
    \label{tracking_error}
    \vspace{-0.5cm}
\end{figure}

We select a representative simulation trial in MuJoCo, where Digit navigates a pedestrian-rich environment. For each walking step, we plot the desired $x$ and $y$ positions and the heading angle, alongside the actual positions and heading angle of the simulated Digit. The tracking results are shown in Fig. \ref{tracking_error}. While the tracking of the $x$-position is highly accurate, the $y$-position tracking error is non-negligible, and the heading angle exhibits more pronounced deviations. As discussed in Eq. (\ref{EqLipModel}), the ROM is primarily designed for sagittal motion and does not explicitly account for body orientation and lateral motion. As a result, $x$-direction tracking remains precise, whereas tracking errors in $y$-position and heading angle become more noticeable, particularly when the desired heading angle changes frequently. Such tracking errors significantly degrade navigation performance when deploying policies trained with ROMs onto full-body bipedal robots, as will be experimentally demonstrated in Section \ref{experiments}-E.

\subsection{Benchmarks}

To highlight the superior social navigation capabilities of our approach, we first disable emotion-related observations and reward functions, ensuring a fair comparison with SOTA baselines. In addition, we adapt the traditional methods, originally designed for wheeled mobile robots, to be deployable on Digit simulated in MuJoCo. To validate the performance discrepancies between the use of full-body bipedal robot dynamics and the use of the ROM, we also deploy certain SOTA approaches that rely exclusively on the ROM. The SOTA baselines are listed below. 

\textbf{DWA}. The Dynamic Window Approach (DWA) \cite{FoxDynamic} is a traditional non-learning method widely applied for local navigation using wheeled mobile robots. In this study, we modify DWA by replacing the kinematics of wheeled mobile robots with the LIP model (\ref{EqLipModel}), enabling optimization of the sagittal speed and heading angle.

\textbf{LNDNL}. The ``Learn to Navigate in Dynamic environments with Normalized LiDAR scans'' (LNDNL) method \cite{ZhuLearn}  is a DRL-based navigation approach that directly maps the raw LiDAR scans to robot actions. LSTM networks are used to process the sequential LiDAR scans, capturing spatio-temporal interaction features.

\textbf{LiDAR--SAC}. LiDAR–SAC is inspired by the SOTA study in~\cite{JinMapless}, using an encoder to extract latent features from sequential LiDAR scans stacked over the time horizon. However, the original approach faces challenges in learning a feasible navigation policy, as empirically validated in \cite{ZhuLearn}. Therefore, we replace the original networks with ours, including the encoder and the SAC structure, to validate the effectiveness of our network pipeline.

\textbf{DRL--VO}. As discussed in Section \ref{relatedWork}, the DRL–VO pipeline \cite{XieLearning} combines fully observable states, such as pedestrian positions and velocities, with observations of raw sequential LiDAR scans. The combined information is transformed into an image, enabling the extraction of features from dynamic and interactive environments using CNNs. However, the original network pipeline consistently takes suboptimal actions, leading to minimal reward. To address this, we improve it by integrating our SAC structure, resulting in the modified DRL–VO*, which will be used as a baseline in our study.

\textbf{SARL and RGL}. SARL \cite{ChenCrowd} and RGL pipelines \cite{ChenRelational} assume fully observable environments. In addition, ego-agents are assumed to have similar dynamics and constraints to pedestrians. However, when the ego-agent is a bipedal robot with velocity constraints in Eq.~\ref{EqAction}, neither SARL nor RGL succeed in achieving a feasible navigation policy, whether using ROMs or full-order dynamics.

\textbf{Single--LGM}. Single--LGM uses only one single LiDAR grid map (LGM) at the current walking step instead of a sequence of historical ones, as illustrated in Fig. \ref{framework}-(a). 

\textbf{OGM}. OGM leverages a sequence of historical occupancy grid maps (OGMs) shown in Fig. \ref{GridMap}-(c) in the frame of the ego-agent, replacing the LGMs illustrated in Fig. \ref{GridMap}-(b).

\textbf{Evaluation Metrics}. We evaluate each benchmark method over 500 random trials. 
The initial positions of dynamic and static objects are set randomly for each evaluation, while the initial and goal positions of Digit's torso remain the same. For each trial, all benchmarks start with the same initial environments, ensuring that dynamic and static objects have the same initial positions and pedestrian emotions are consistent. The success rate (\texttt{SR}) is defined as the ratio of successful episodes that occur without collision or timeout. The navigation time (\texttt{NT}) represents the average time taken for all successful episodes. Discomfort times (\texttt{DT}) count the number of walking steps when Digit enters the discomfort zone of a static obstacle or a pedestrian.

\begin{figure}[!t]
    \centering
    \small
    \subfigure[Training process]{
	\includegraphics[height=1.45in]{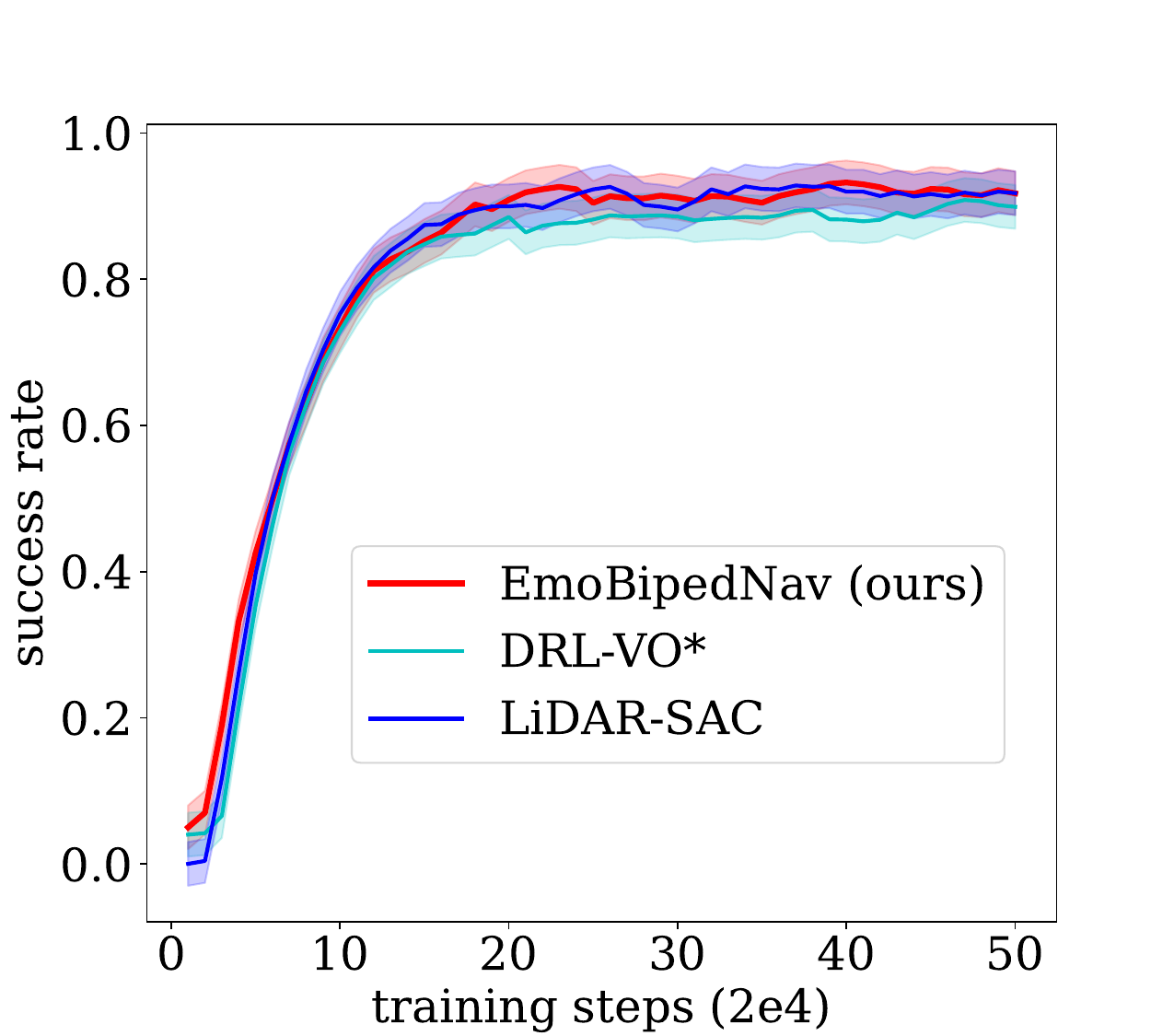} 
    }
    \subfigure[Training scene in Mujoco]{
	\includegraphics[height=1.45in]{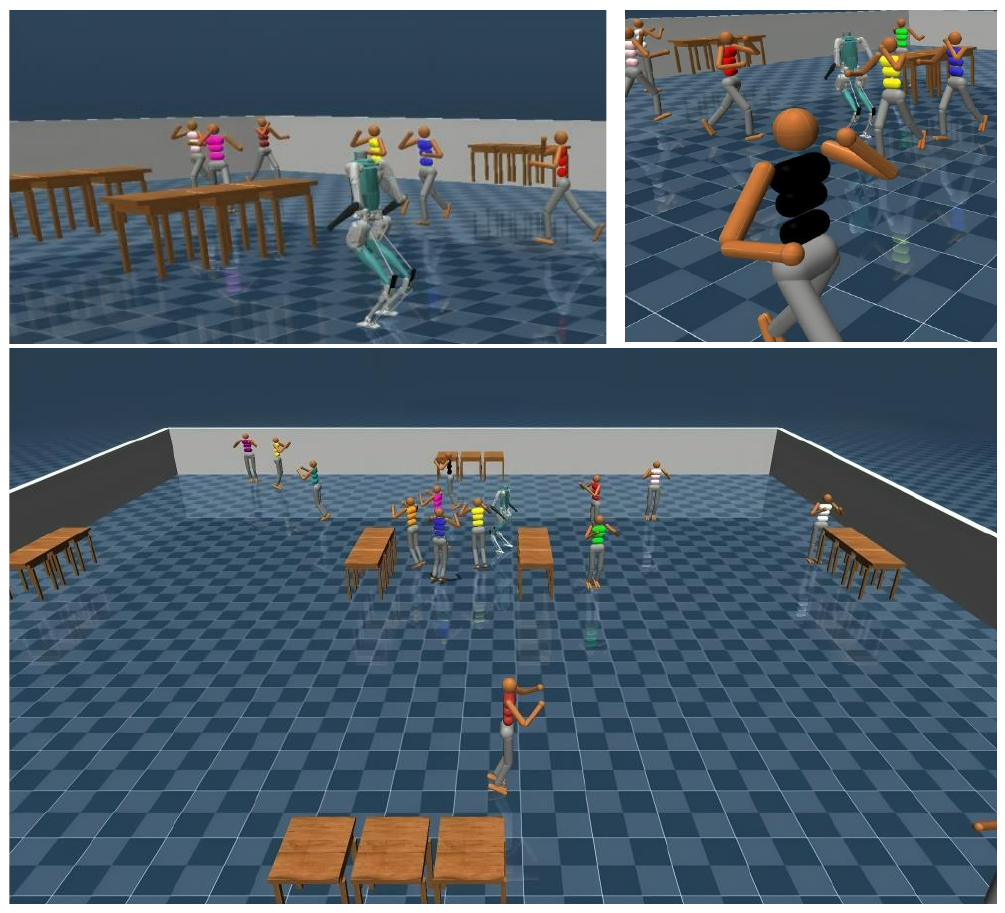} 
    }
    \DeclareGraphicsExtensions.
    \caption{Training process with simulated Digit in MuJoCo. (a) shows the training processes of different approaches and (b) illustrates a representative simulation scene. We evaluate each model 100 times in randomly generated environments at every $2 \times 10^4$ training steps. The success rate is calculated based on these 100 randomized evaluations. DRL--VO* and LiDAR-SAC are enhanced by replacing their original network frameworks with our proposed architecture. LNDNL is omitted because its success rate climbs to 0.66 and decreases to 0. In addition, SARL and RGL are excluded from the analysis as they fail. The trainings of Single--LGM and OGM are removed due to their similar learning curves to our method.}
    \label{training}
\end{figure}

\renewcommand\arraystretch{1.0}
\begin{table}[!t]
	\small
	\centering
	\caption{Final performance evaluation on the full-body bipedal robot with 500 random tests.}
	\begin{tabular}{c c c c} 
		\hline
		Method           & \texttt{SR}               & \texttt{NT}                    & \texttt{DT} \\
		\hline 
		DWA              & 0.892            & \textbf{27.20}$\pm$3.55  & 415\\
            LNDNL            & 0.660            & 38.13$\pm$5.99           & 301\\
		DRL--VO*         & 0.912            & 28.16$\pm$\textbf{2.49}  & 318\\
            LiDAR--SAC       & 0.932            & 27.24$\pm$3.10           & 766\\
            Single--LGM      & 0.928            & 33.15$\pm$4.84           & 546\\
            OGM              & 0.944            & 29.45$\pm$3.79           & 312\\
		EmoBipedNav (ours)    & \textbf{0.950}   & 28.83$\pm$3.15           & \textbf{226}\\
		\hline
	\end{tabular}
	\label{tableComparisonMujoco}
\end{table}

As illustrated in Fig. \ref{training}, both LiDAR--SAC and DRL--VO* produce results similar to ours when their original network structures are replaced with the architecture described in Section \ref{approach}, further validating the robustness and effectiveness of our network pipeline. Moreover, DRL-VO* has the same network structure as ours but yields inferior performance in terms of the stable success rate, further indicating that the proposed LGMs are better representations. 

The quantitative results, presented in TABLE \ref{tableComparisonMujoco}, confirm that our method achieves the highest \texttt{SR} for the deployment on the full-body bipedal robot Digit. DWA enables efficient navigation with the shortest average time of 27.2 s, at the cost of decreasing \texttt{SR} (0.892) and increasing \texttt{DT}. It is noteworthy that DWA can reach a higher \texttt{SR} (0.950) when deployed with the LIP model without considering full-order dynamics, which underscores the notable dynamics discrepancies between the ROM and the full-body robot dynamics. LNDNL demonstrates significantly inferior performance in terms of both \texttt{SR} and \texttt{NT} due to the plateauing problem during training. Notably, we observed that LNDNL achieves results comparable to our approach when trained and evaluated with the ROM without incorporating full-body dynamics, emphasizing our pipeline's capability to effectively handle the complex dynamics of the full-body robot. Although Single--LGM can also achieve a high \texttt{SR}, its \texttt{NT} significantly increases due to its conservative policy, which does not account for the temporal dynamics of the social environment. OGM, on the other hand, shows a comparable \texttt{SR}, but its \texttt{DT} is higher than our method, indicating that LGMs are more effective in reducing discomfort.

\subsection{Emotion-integrated Implementation}

\begin{figure}[!t]
    \centering
    \small
    \subfigure[Different discomfort distances]{
	\includegraphics[width=1.6in]{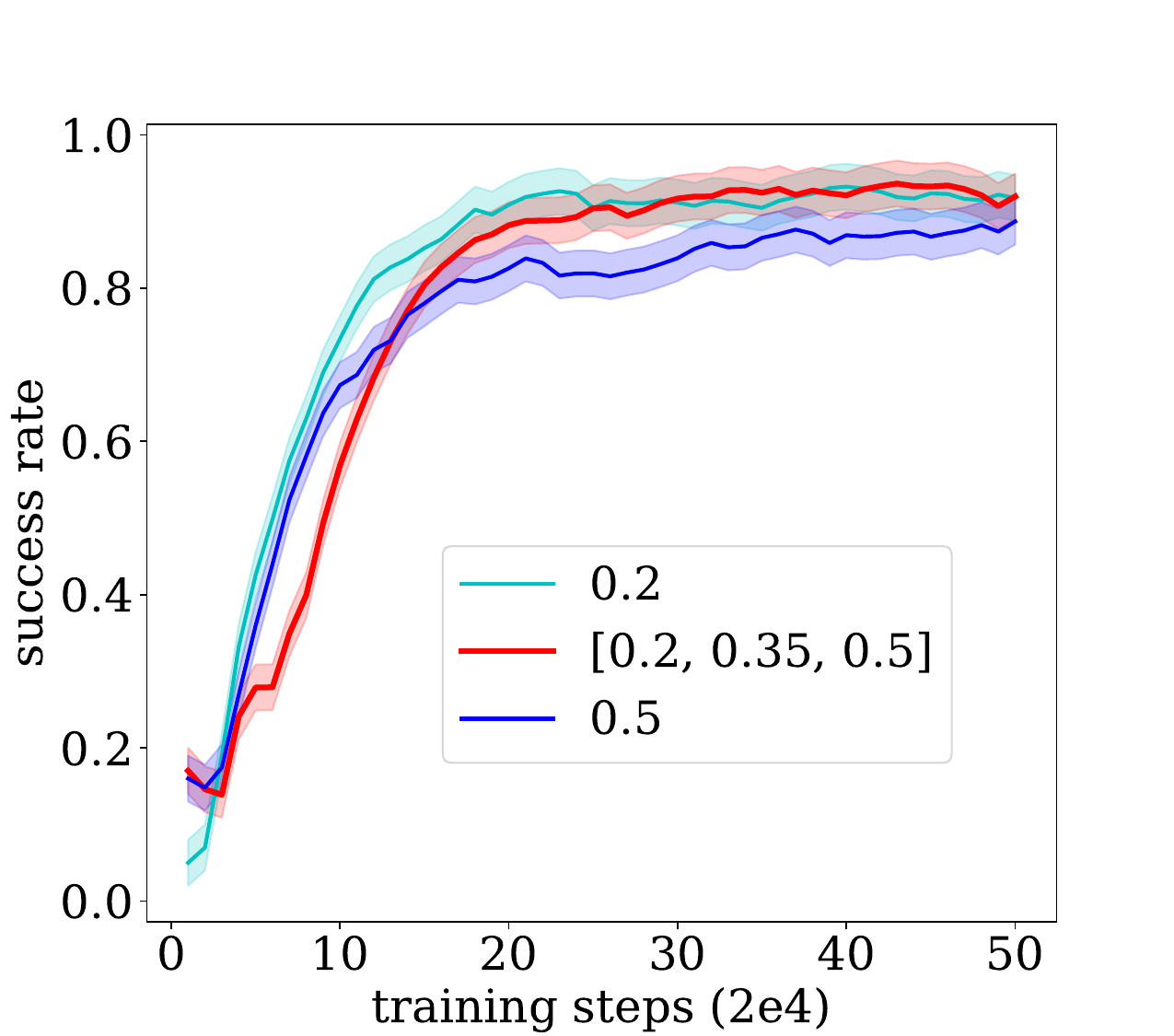} 
    }
    \subfigure[\centering Different observation and model]{
	\includegraphics[width=1.6in]{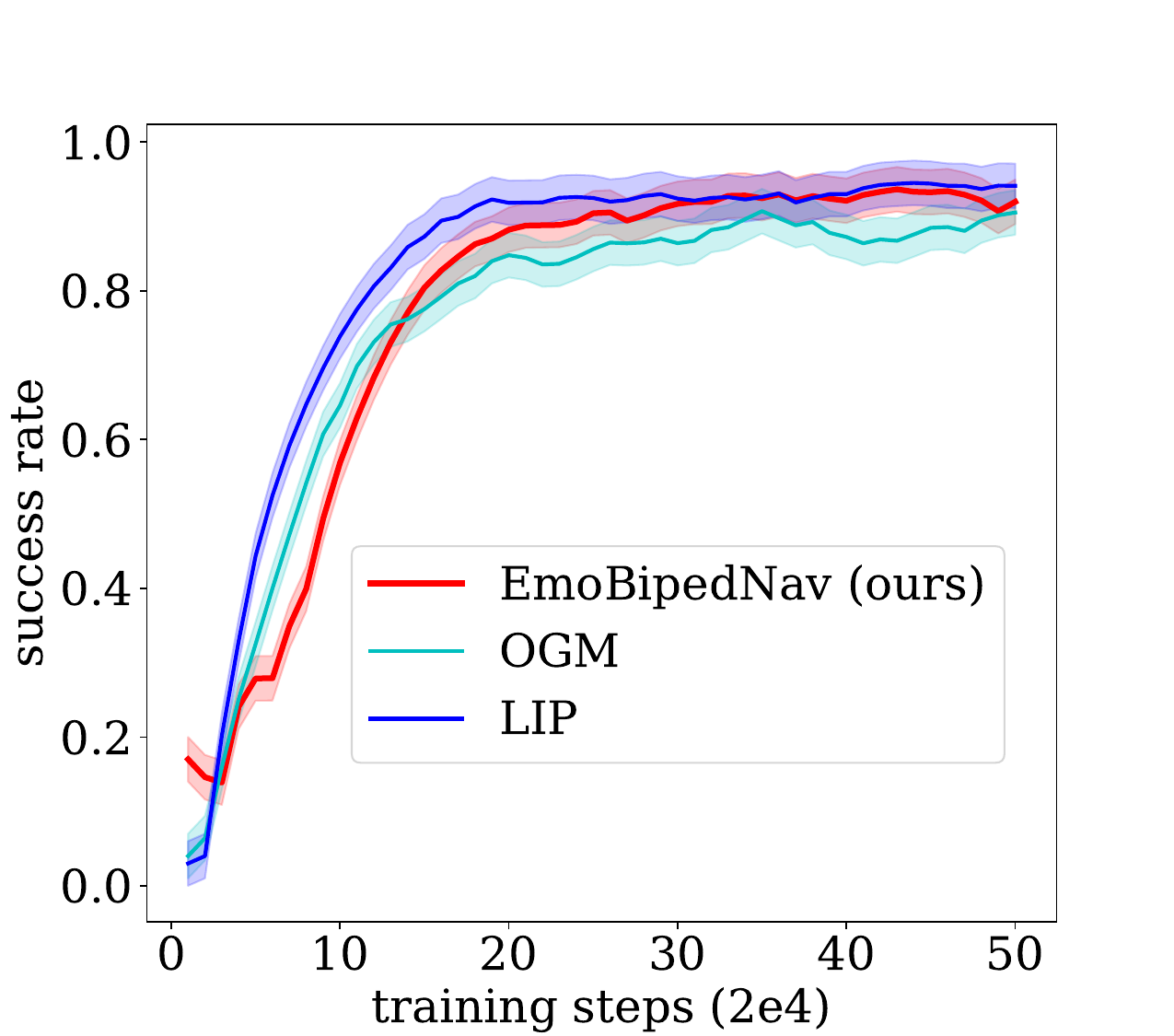} 
    }
    \DeclareGraphicsExtensions.
    \caption{Training conditioned with emotions. (a) illustrates the learning processes using two constant discomfort distances and one variable discomfort distance corresponding to different pedestrian emotions. All trainings in (b) use the variable discomfort distance. EmoBipedNav and OGM leverage full-order dynamics while the LIP one only relies on the ROM.}
    \label{trainingEmotion}
\end{figure}

To investigate further, we conduct an additional training session to assess the performance of integrating emotion observations. The results demonstrate that our approach can adapt to pedestrians with different emotions, showing the emotion-aware feature of our navigation policy. Additionally, we validate the performance discrepancies of robot social navigation when using the ROM and the full-body bipedal robot. Moreover, we practically verify that using LGMs provides greater benefits for collision and discomfort avoidance compared to using OGMs in these emotion-aware environments.

First, we use variable discomfort distances (Var--DD) indicating different emotions of pedestrians, and the learning process is similar to the case where a constant discomfort distance of $0.2$ m is used, as shown in Fig. \ref{trainingEmotion}-(a). This similarity suggests that our approach can effectively avoid collisions in emotion-integrated environments. Next, we set the discomfort distance to a constant $0.5$ m, resulting in a learning process similar to the other two settings. However, progress slows, and the stable success rate is lower. We attribute this to the aggressive pedestrian motion generator ORCA \cite{BergReciprocal}, which challenges the conservative movement of the ego-agent to avoid collisions. Furthermore, we replace LGMs with OGMs, resulting in similar learning performance. However, as shown in Fig. \ref{trainingEmotion}-(b), our approach demonstrates a larger stable success rate, underscoring the advantages of LGMs for collision avoidance. Lastly, we train our pipeline using only the LIP robot model. Although learning based on the LIP model is more efficient due to simplified dynamics, performance with respect to \texttt{SR} and \texttt{DT} significantly degrades when the learned navigation policy is deployed on Digit, as demonstrated in TABLE \ref{tableAblations}.

\renewcommand\arraystretch{1.0}
\begin{table}[!t]
	\small
	\centering
        \caption{Performance evaluation w/o emotions and w/o full-body robot dynamics with 500 random tests.}
	\begin{tabular}{c@{\hspace{2pt}}c@{\hspace{7pt}}c@{\hspace{10pt}}c@{\hspace{10pt}}c} 
		\hline
		DD (Train)       & Train/Test     & Obs    & \texttt{IDT} (Test)   & \texttt{SR}                    \\
		\hline 
            $0.2$ m       & Digit/Digit    & LGM            & [\textbf{51}; 50,      181,           1352]            & 0.950                        \\
            $0.5$ m       & Digit/Digit    & LGM            & [395;         81,      \textbf{138},  \textbf{1167}]   & 0.938               \\
            \rowcolor{gray!20} \multicolumn{5}{l}{~~Var-DD~~~Digit/Digit~~LGM~~~[213;         56,      166,           1198]~~~~~~0.940} \\
            \hline
            Var-DD    & Digit/Digit    & OGM            & [436;         87,      160,           1174]            & 0.892                       \\
            Var-DD    & LIP/LIP        & LGM            & [202; \textbf{37},     143,           1227]            & \textbf{0.962}               \\
            Var-DD    & LIP/Digit      & LGM            & [642;          66,     177,           1266]            & 0.900                        \\
		\hline
            \multicolumn{5}{c}{DD: discomfort distance (m); Obs: observation}\\
		\multicolumn{5}{c}{Var-DD$ = [0.2, 0.35, 0.5]$ m $\leftrightarrow$ [\texttt{happy}, \texttt{neutral}, \texttt{negative}]}\\
		\hline
	\end{tabular}
	\label{tableAblations}
\end{table}

To account for pedestrian emotions and their associated discomfort distances, we introduce an additional evaluation metric, \texttt{IDT}, which quantifies the discomfort times for individual entities. Specifically, the four values in the \texttt{IDT} column of TABLE \ref{tableAblations} correspond to the \texttt{IDT} of static obstacles and pedestrians exhibiting \texttt{happy}, \texttt{neutral}, and \texttt{negative} emotions, respectively. During training, the two baseline models enforce a fixed discomfort distance of $0.2$ m or $0.5$ m for all dynamic pedestrians, regardless of their emotional states. However, during testing, \texttt{IDT} is computed based on variable discomfort distances of $[0.2, 0.35, 0.5]$ m, corresponding to the three emotional states. This discrepancy between training and testing conditions allows us to evaluate the model's ability to adapt to varying discomfort zones.

Compared to maintaining a constant discomfort distance of $0.2$ m, our approach of using variable discomfort distances of $[0.2, 0.35, 0.5]$ m decreases \texttt{IDT} for \texttt{neutral} and \texttt{negative} pedestrians while increasing  \texttt{IDT} for \texttt{happy} pedestrians, and static obstacles in particular. This indicates that our approach prioritizes \texttt{neutral} and \texttt{negative} pedestrians by assigning them longer discomfort distances. However, this comes at the cost of reduced comfort for \texttt{happy} pedestrians and, notably, static obstacles. The performance degradation with respect to static obstacles may result from the reduction in the available free areas. This occurs because the discomfort zones of \texttt{neutral} and \texttt{negative} pedestrians can expand to occupy significantly more space while the overall environment space remains the same. Furthermore, the ego-agent is inherently more prone to collisions with moving pedestrians due to highly dynamic interactions. As a result, it prioritizes maintaining distance from these pedestrians, and inadvertently increases the likelihood of intruding into the discomfort zones of static obstacles. When the constant discomfort distance is increased to $0.5$ m, the ego-agent further improves \texttt{IDT} for \texttt{neutral} and \texttt{negative} pedestrians, yet compromising comfort for \texttt{happy} pedestrians and static obstacles. The increased \texttt{IDT} for static obstacles could be attributed to the same factors observed when using variable discomfort distances $[0.2, 0.35, 0.5]$ compared to a fixed discomfort distance of $0.2$ m. In addition, the larger discomfort distance leads to a decline in \texttt{SR}, which may stem from the aggressive pedestrian motion generated by ORCA \cite{BergReciprocal}, despite the conservative behaviors of the ego-agent for intrusion and collision avoidance.

When replacing our LGMs with OGMs, both \texttt{SR} and \texttt{IDT} of static obstacles significantly degrade, further demonstrating the effectiveness of LGM in enhancing collision and discomfort avoidance. Additionally, while the pipeline trained and tested with the LIP model achieves the highest overall performance, its \texttt{SR} drops to 0.9, and \texttt{IDT} increases substantially when tested with Digit, further highlighting the limitations of ROMs in maintaining robust navigation performance.

\subsection{Implementation with Time-varying Emotions}
\begin{figure}[!t]
    \centering
    \small
    \subfigure[Fixed emotion]{
	\includegraphics[width=1.5in]{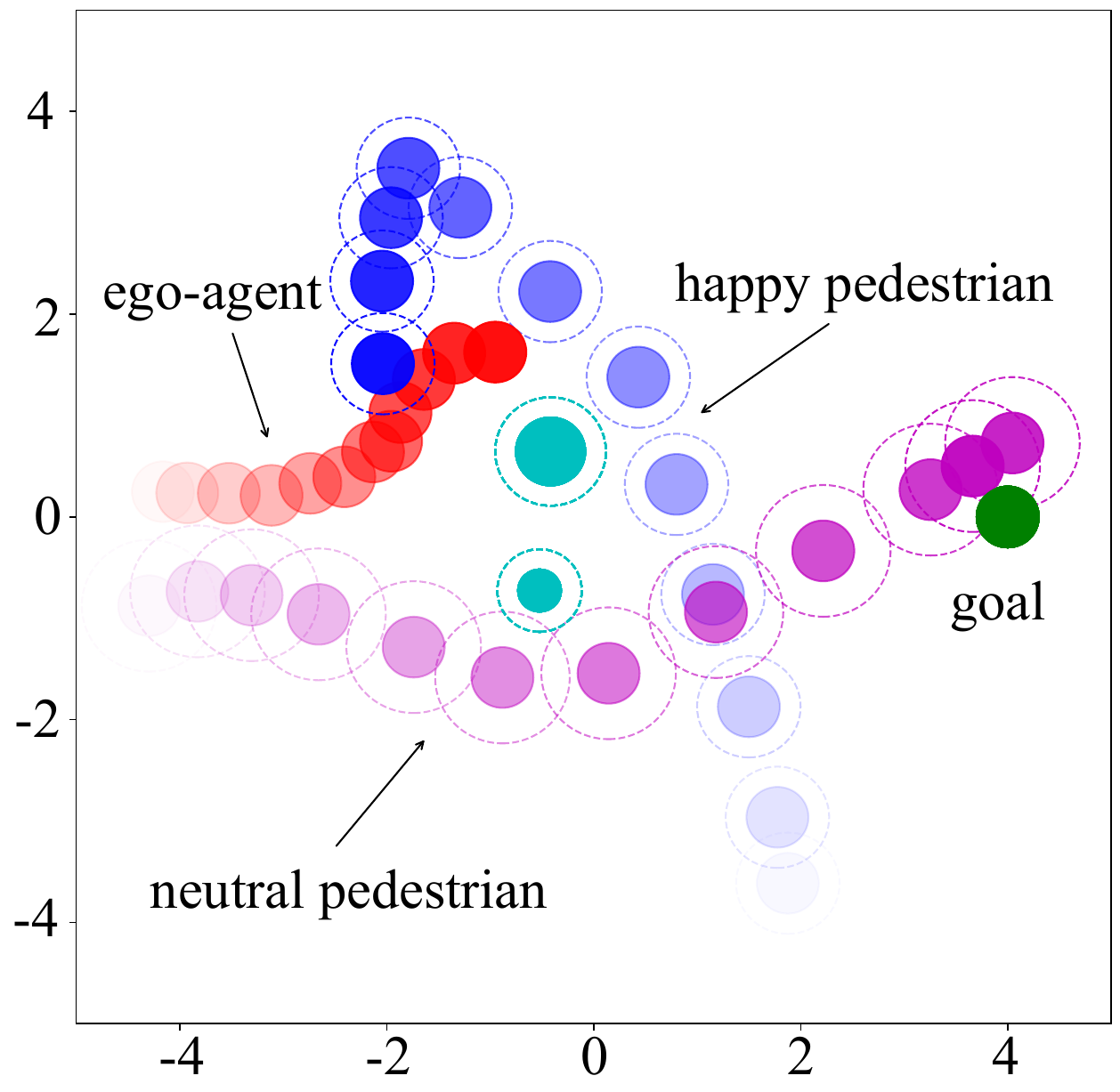} 
    }
    \subfigure[Time-varying emotion]{
	\includegraphics[width=1.5in]{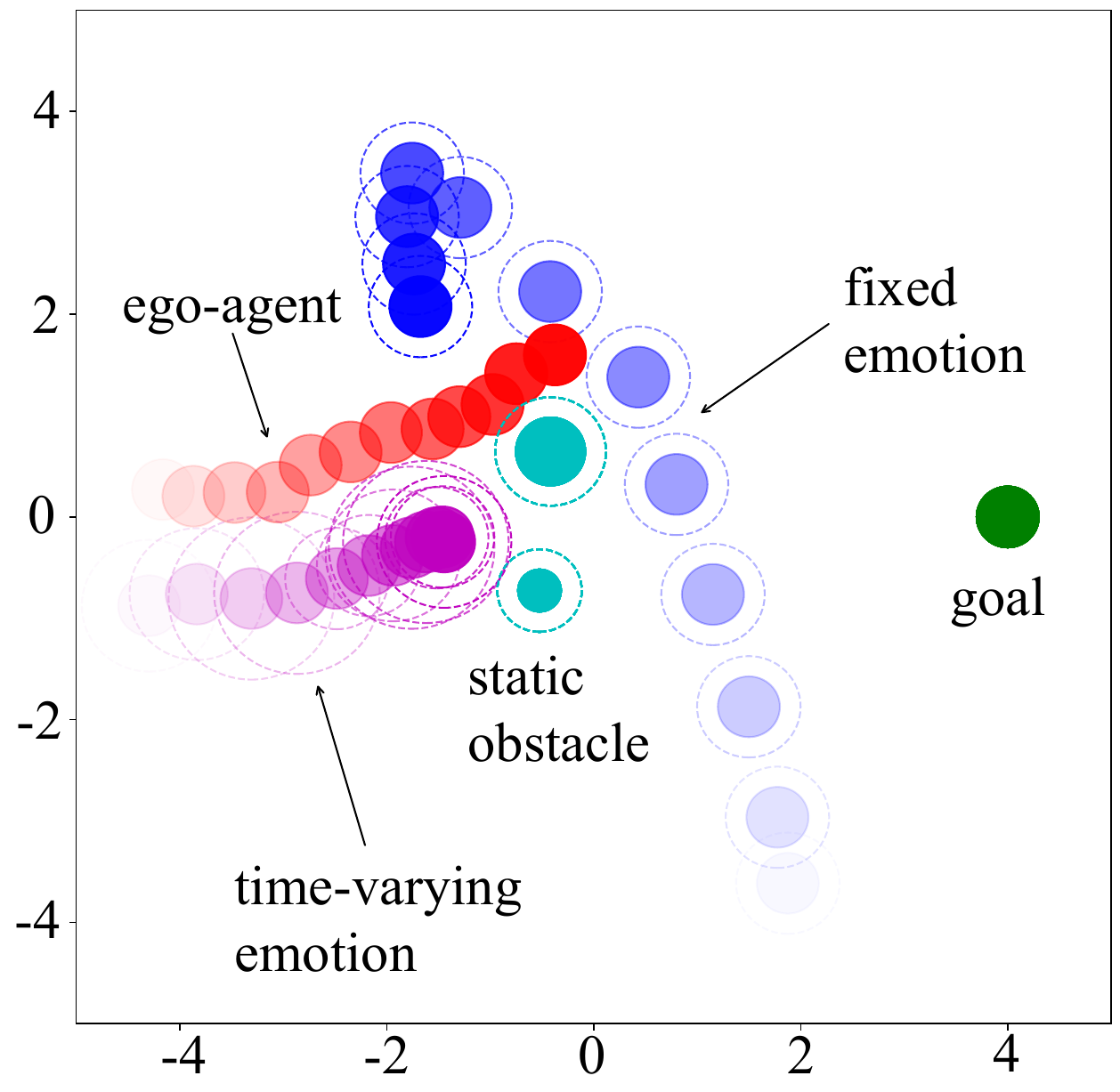} 
    }
    \DeclareGraphicsExtensions.
    \caption{Robot and pedestrian trajectories under different emotion settings. (a) The emotions of two pedestrians remain fixed, with one classified as \texttt{happy} and the other as \texttt{neutral}. (b) One pedestrian's emotion dynamically changes every 2 seconds. The decreasing transparency illustrates the time evolvement.}
    \label{robustnessVerification}
\end{figure}

In real-world scenarios, emotions can fluctuate or be misclassified due to inaccuracy in the facial recognition algorithm, which is a pretrained CNN model relying on small dataset \footnote{\label{emotionNet}https://github.com/susantabiswas/realtime-facial-emotion-analyzer}. To investigate the adaptability of our approach to these uncertainties, we introduce a more challenging simulation setup where one pedestrian's emotion keeps alternating among \texttt{happy}, \texttt{neutral}, and \texttt{negative} every 2 seconds. Statistically, \texttt{SR} roughly remains the same, while \texttt{IDT} shifts from $[213; 56, 166, 1198]$ to $[258; 42, 204, 1317]$. The similar \texttt{SR} shows that our approach is stable for collision avoidance, although pedestrian emotions vary over time. The increased \texttt{IDT} for \texttt{neutral} and \texttt{negative} pedestrians may result from the ego-agent intruding into larger discomfort zones when nearby pedestrians suddenly change to a \texttt{negative} emotional state from \texttt{neutral}, or from \texttt{happy} to \texttt{neutral}. Fig. \ref{robustnessVerification} presents a representative episode, comparing the case with fixed emotions (Fig. \ref{robustnessVerification}-(a)) to that with time-varying emotions (Fig. \ref{robustnessVerification}-(b)). The latter scenario shows increased discomfort times, particularly at the beginning of the episode when a nearby pedestrian frequently changes the emotion, particularly from \texttt{neutral} to \texttt{negative}. Moreover, frequently changing pedestrian emotion not only impacts the ego-agent's trajectory, but also alters pedestrians' interactions, indicating complex real-time interactions between robot-pedestrian, further verifying the adaptability of our social navigation policy.

\subsection{Hardware Demonstrations}
\begin{figure*}[!t]
	\centering
	\small
	\includegraphics[width=17.0cm]{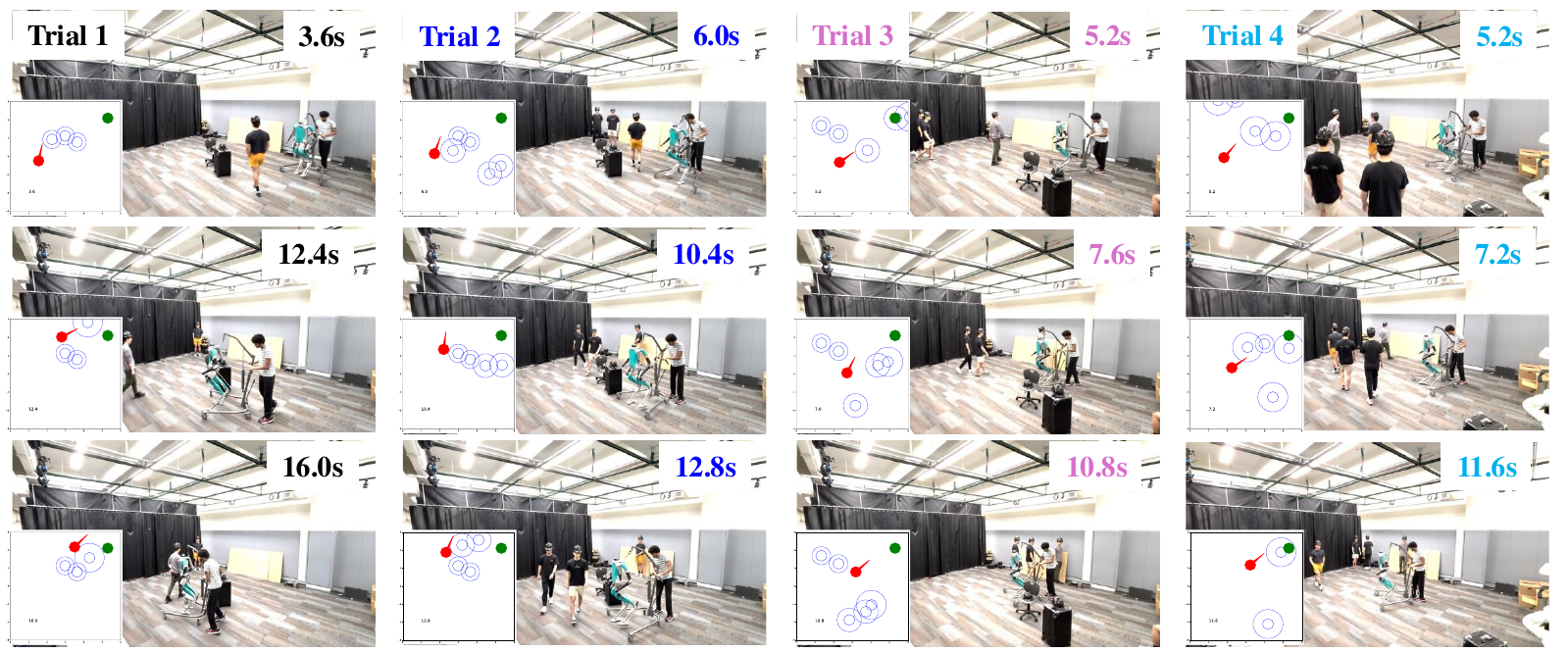}
	\caption{Screenshots of hardware deployments under various pedestrian navigation patterns. Each column corresponds to an implementation. In each screenshot, a LiDAR scan figure is overlaid to illustrate the robot's position and orientation, represented by a solid red circle and a red arrow, respectively. Object collision margins are depicted as blue hollow circles, while discomfort margins are indicated with dashed circles. We track the robot and pedestrians with an external motion capture system for accurate trajectory analysis. Pedestrian emotions are randomly set at the beginning and remain constant for each implementation.}
	\label{MocapScreenshot}
\end{figure*}

\begin{figure*}[!t]
	\centering
	\small
	\includegraphics[width=17.0cm]{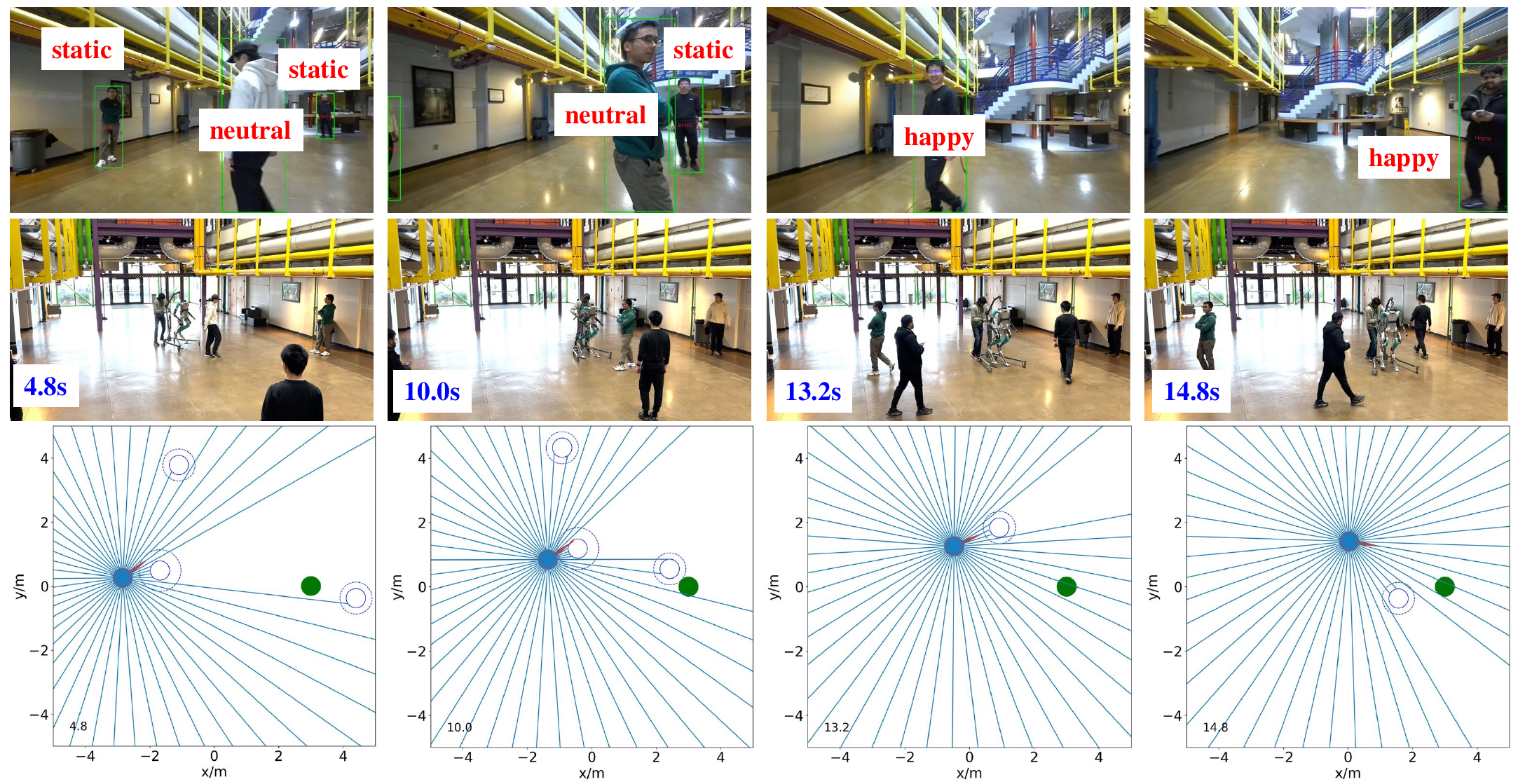}
	\caption{Emotion recognition with an on-board ZED camera. The top row presents the ZED camera's perspective, showcasing detected pedestrians along with their recognized emotions. The middle row shows the recorded scenes from a fixed external camera view. The bottom figures illustrate a top-down view of the reconstructed LiDAR scan, position and orientation of the ego-agent, and collision and discomfort margins of pedestrians.}
	\label{RealScreenshot}
\end{figure*}

\textbf{LiDAR scan reconstruction}. Since real-world objects have irregular geometries, whereas simulated objects are assumed to be circular, directly using raw LiDAR scans in real environments leads to sensor mismatches, creating sim-to-real observation gaps for our navigation policy. To mitigate this, we reconstruct LiDAR scans to better align with the simulated environment. Specifically, we first determine the positions of surrounding objects in the ego-agent frame and approximate their shapes as circles with a fixed radius of $0.3$ m for simplification. Next, we emit $K$ rays from the ego-agent to detect these circular approximations, generating a reconstructed LiDAR scan that more closely resembles the simulated observations.

\textbf{Deployments under various pedestrian motion patterns}. We begin by transferring the simulated policy learned in MuJoCo \cite{TodorovMujoco} to the physical robot Digit. Furthermore, we include various pedestrian motion patterns to verify the sim-to-real capabilities of our emotion-aware navigation policy. Fig. \ref{MocapScreenshot} shows four real-world implementations. The first three columns depict scenarios involving dynamic pedestrians and static obstacles, while the last row showcases an environment where all pedestrians are in dynamic movement. Notably, in the first column, a pedestrian intentionally walks in front of Digit and remains stationary to obstruct its path. Additionally, pedestrians move ingroup together in the second and third columns. Despite these diverse and challenging conditions, Digit successfully reaches its goal without collision while actively avoiding intrusions into discomfort zones.

\textbf{Deployments integrating emotion recognition}. To further validate the practicality of our pipeline with integrated pedestrian emotion recognition, we equip the physical Digit with a ZED stereo camera. Robot localization and pedestrian detection are performed using camera APIs \footnote{https://www.stereolabs.com/docs}. For each detected pedestrian, facial features are extracted using OpenCV libraries \footnote{https://docs.opencv.org/3.4/db/d28/tutorial$_-$cascade$_-$classifier.html}, and emotions are recognized using a pre-trained model \footref{emotionNet}. If no face is detected -- due to factors such as long distance, occlusion by other pedestrians, or the pedestrian facing away from the camera -- the emotion is set to \texttt{neutral}. Localization and pedestrian detection are performed on a Jetson Orin at 10 Hz, while face extraction and emotion recognition operate at 0.5 Hz. Meanwhile, the low-level controller and social navigation policy run on an Intel NUC at 1000 Hz and 2.5 Hz, respectively.

Real-world deployments, as demonstrated in the accompanying video, confirm the seamless transferability of our navigation framework from simulation to real-world applications. Fig. \ref{RealScreenshot} provides representative screenshots from a hardware demonstration. When pedestrians are far from Digit or facing away from the camera, their faces cannot be detected and are therefore labeled with \texttt{neutral} emotions. Once their faces become clearly visible, their emotions change to \texttt{happy}. In the accompanying video, we showcase additional scenarios where pedestrian emotions transition from \texttt{neutral} to \texttt{negative} or shift between other emotional states. Furthermore, we operate Digit on campus with more realistic social interactions, further demonstrating the practicality of our emotion-aware social navigation pipeline for bipedal robots (see the attached video).

\section{DISCUSSION}
In this study, we present a DRL-based navigation framework designed to enable bipedal robots to navigate socially interactive environments integrated with pedestrian emotions. Future research will focus on incorporating more realistic social norms, such as adhering to conventions like ``walking on the right'', and conducting DRL training in increasingly complex social scenarios, including walls and grouped pedestrians. The ultimate goal is to facilitate natural, socially compliant, and effective locomotion for bipedal robots in crowded and dynamic environments.


\vspace{-14mm}
\begin{IEEEbiography}[{\includegraphics[width=1in,height=1.25in,clip,keepaspectratio]{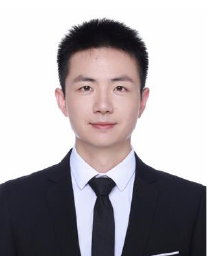}}]{Wei Zhu} (Member, IEEE) received the Bachelor and Master degrees from Nankai University, Tianjin, China, in 2017 and 2020, respectively, and the Ph.D. from Tohoku University, Sendai, Japan, in 2023. He is a Post-Doctoral Scholar at Georgia Institute of Technology, Atlanta, GA, USA. He will join Tohoku University as an Assistant Professor in 2025. His research interests include deep reinforcement learning, robot navigation in crowds, bipedal robots, and snake-like robots.
\end{IEEEbiography}

\vspace{-14mm} 

\begin{IEEEbiography}[{\includegraphics[width=1in,height=1.25in,clip,keepaspectratio]{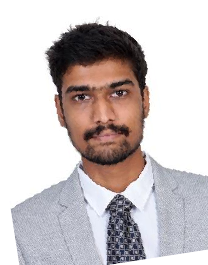}}]{Abirath Raju} received the B.Tech degree in mechanical engineering with a minor in computer science from the National Institute of Technology, Tiruchirappalli, Tamil Nadu, India, in 2023. He is currently pursuing the M.S. degree in robotics at Georgia Institute of Technology, Atlanta, GA, USA. His research interests include bipedal robot navigation and planning, computer vision and robot learning.
\end{IEEEbiography}

\vspace{-14mm}

\begin{IEEEbiography}[{\includegraphics[width=1in,height=1.25in,clip,keepaspectratio]{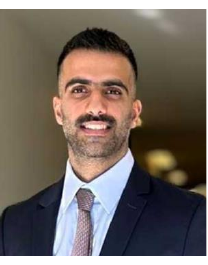}}]{Abdulaziz Shamsah} received the B.S. degree in mechanical engineering from the Rensselaer Polytechnic Institute, Troy, NY, USA, in 2016, and the M.S.E. degree in mechanical engineering and applied mechanics from the University of Pennsylvania, Philadelphia, PA, USA, in 2018., and the Ph.D. degree in mechanical engineering from the Georgia Institute of Technology, Atlanta, GA, USA, in 2024. He is currently an Assistant Professor in the Mechanical Engineering Department at the College of Engineering and Petroleum, Kuwait University, Kuwait.  His research interests include bipedal navigation in real-world environments, formal methods, and safety and robustness in locomotion.
\end{IEEEbiography}

\vspace{-14mm}

\begin{IEEEbiography}[{\includegraphics[width=1in,height=1.25in,clip,keepaspectratio]{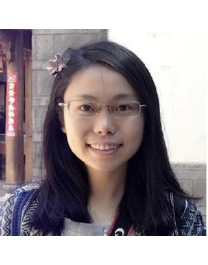}}]{Anqi Wu} is an Assistant Professor at the School of Computational Science and Engineering (CSE), Georgia Institute of Technology. She was a Postdoctoral Research Fellow at the Center for Theoretical Neuroscience, the Zuckerman Mind Brain Behavior Institute, Columbia University. She received her Ph.D. degree in Computational and Quantitative Neuroscience and a graduate certificate in Statistics and Machine Learning from Princeton University. Anqi was selected for the MIT Rising Star in EECS, DARPA Riser, Alfred P. Sloan Fellow, and Kavli Fellow by National Academy of Sciences. Her research focuses on developing scientifically grounded statistical models to uncover structure in neural and behavioral data at the intersection of machine learning and computational neuroscience. She is broadly interested in creating data-driven models to advance biological intelligence and artificial intelligence.
\end{IEEEbiography}

\vspace{-14mm}

\begin{IEEEbiography}[{\includegraphics[width=1in,height=1.25in,clip,keepaspectratio]{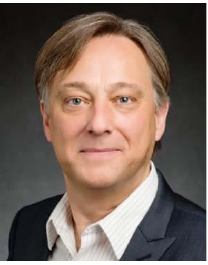}}]{Seth Hutchinson} (Fellow, IEEE) received the Ph.D. degree in electrical engineering from Purdue University, West Lafayette, IN, USA, in 1988.

Seth Hutchinson is a professor at Northeastern University. He was previously the Executive Director of the Institute for Robotics and Intelligent Machines at the Georgia Institute of Technology, where he was also Professor and KUKA Chair for Robotics in the School of Interactive Computing (2018-2024), and he is Professor emeritus in the ECE Department at the University of Illinois in Urbana-Champaign, where he was a faculty member during 1990-2017. He received his Ph.D. from Purdue University. 

Hutchinson served as president of the IEEE Robotics and Automation Society, Editor-in-Chief for the ``IEEE Trans. on Robotics'' and was founding Editor-in-Chief of the RAS Conference Editorial Board.  He has more than 300 publications on the topics of robotics and computer vision, and is coauthor of two books on robotics.
\end{IEEEbiography}

\vspace{-14mm}

\begin{IEEEbiography}[{\includegraphics[width=1in,height=1.25in,clip,keepaspectratio]{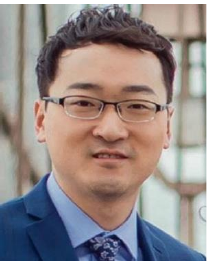}}]{Ye Zhao}  (Senior Member, IEEE) received the Ph.D. degree in mechanical engineering from The University of Texas at Austin in 2016. He was a Post-Doctoral Fellow with the John A. Paulson School of Engineering and Applied Sciences, Harvard University, Cambridge, MA, USA. He is currently an Assistant Professor at George W. Woodruff School of Mechanical Engineering, Georgia Institute of Technology, Atlanta, GA, USA. His research interests include robust task and motion planning, contact-rich trajectory optimization, and formal methods for legged locomotion and navigation.

Dr. Zhao received the CAREER Award from the National Science Foundation in 2022 and the Young Investigator Award from the Office of Naval Research in 2023. He serves as an Associate Editor for IEEE Transactions on Robotics, IEEE/ASME Transactions on Mechatronics, IEEE Robotics and Automation Letters, and IEEE Control Systems Letters.
\end{IEEEbiography}

\end{document}